\documentclass[10pt,twocolumn,letterpaper]{article}
\usepackage[accsupp]{axessibility} 
\usepackage{wacv} %
\usepackage{graphicx}
\usepackage{amsmath}
\usepackage{amssymb}
\usepackage{booktabs}
\usepackage{graphicx}
\usepackage{booktabs}
\usepackage{float}
\usepackage{footnote}
\usepackage{multirow}
\usepackage{siunitx,etoolbox}
\usepackage{amssymb}
\usepackage{pifont}
\newcommand{\cmark}{\ding{51}}
\newcommand{\xmark}{\ding{55}}
\usepackage[table]{xcolor}
\usepackage{wrapfig}
\usepackage{listings}
\usepackage{enumitem}
\usepackage{svg}
\usepackage{makecell}

\DeclareMathOperator*{\argmin}{arg\,min}

\newcommand{\mysubsubsection}[1]{
  \phantomsection
  \vspace{0.5\baselineskip}
  \vspace{\parskip}
  \noindent\textbf{#1}
}

\usepackage[pagebackref,breaklinks,colorlinks]{hyperref}

\usepackage[capitalize]{cleveref}
\crefname{section}{Sec.}{Secs.}
\Crefname{section}{Section}{Sections}
\Crefname{table}{Table}{Tables}
\crefname{table}{Tab.}{Tabs.}
\def\wacvPaperID{110} 
\def\confName{WACV}
\def\confYear{2025}

\begin{document}

\title{Enhancing Monocular Depth Estimation with Multi-Source Auxiliary Tasks}

\author{Alessio Quercia $^{1,4}$ \quad
Erenus Yildiz $^{1}$ \quad
Zhuo Cao $^{1}$ \quad
Kai Krajsek $^{3}$ \\ \newline
Abigail Morrison $^{2,4}$ \quad
Ira Assent $^{1,5}$ \quad
Hanno Scharr $^{1}$ \\ \newline
{$^{1}$ IAS-8 $^{2}$ IAS-6 $^{3}$ JSC, Forschungszentrum Juelich, Juelich, Germany } \\
{$^{4}$ Dept. of Computer Science, RWTH Aachen University, Aachen, Germany} \\
{$^{5}$ Dept. of Computer Science, Aarhus University, Aarhus, Denmark} \\ 
{\tt\small\{a.quercia,e.yildiz,z.cao,k.krajsek,a.morrison,i.assent,h.scharr\}@fz-juelich.de}
}

\maketitle

\begin{abstract}
Monocular depth estimation (MDE) is a challenging task in computer vision, often hindered by the cost and scarcity of high-quality labeled datasets. We tackle this challenge using auxiliary datasets from related vision tasks for an alternating training scheme with a shared decoder built on top of a pre-trained vision foundation model, while giving a higher weight to MDE. Through extensive experiments we demonstrate the benefits of incorporating various in-domain auxiliary datasets and tasks to improve MDE quality on average by $\sim11\%$. Our experimental analysis shows that auxiliary tasks have different impacts, confirming the importance of task selection, highlighting that quality gains are not achieved by merely adding data. Remarkably, our study reveals that using semantic segmentation datasets as Multi-Label Dense Classification (MLDC) often results in additional quality gains. Lastly, our method significantly improves the data efficiency for the considered MDE datasets, enhancing their quality while reducing their size by at least 80\%. This paves the way for using auxiliary data from related tasks to improve MDE quality despite limited availability of high-quality labeled data. Code is available at {\small\url{https://jugit.fz-juelich.de/ias-8/mdeaux}}.
\end{abstract}

\begin{figure}[htb]
\centering
\resizebox{\columnwidth}{!}{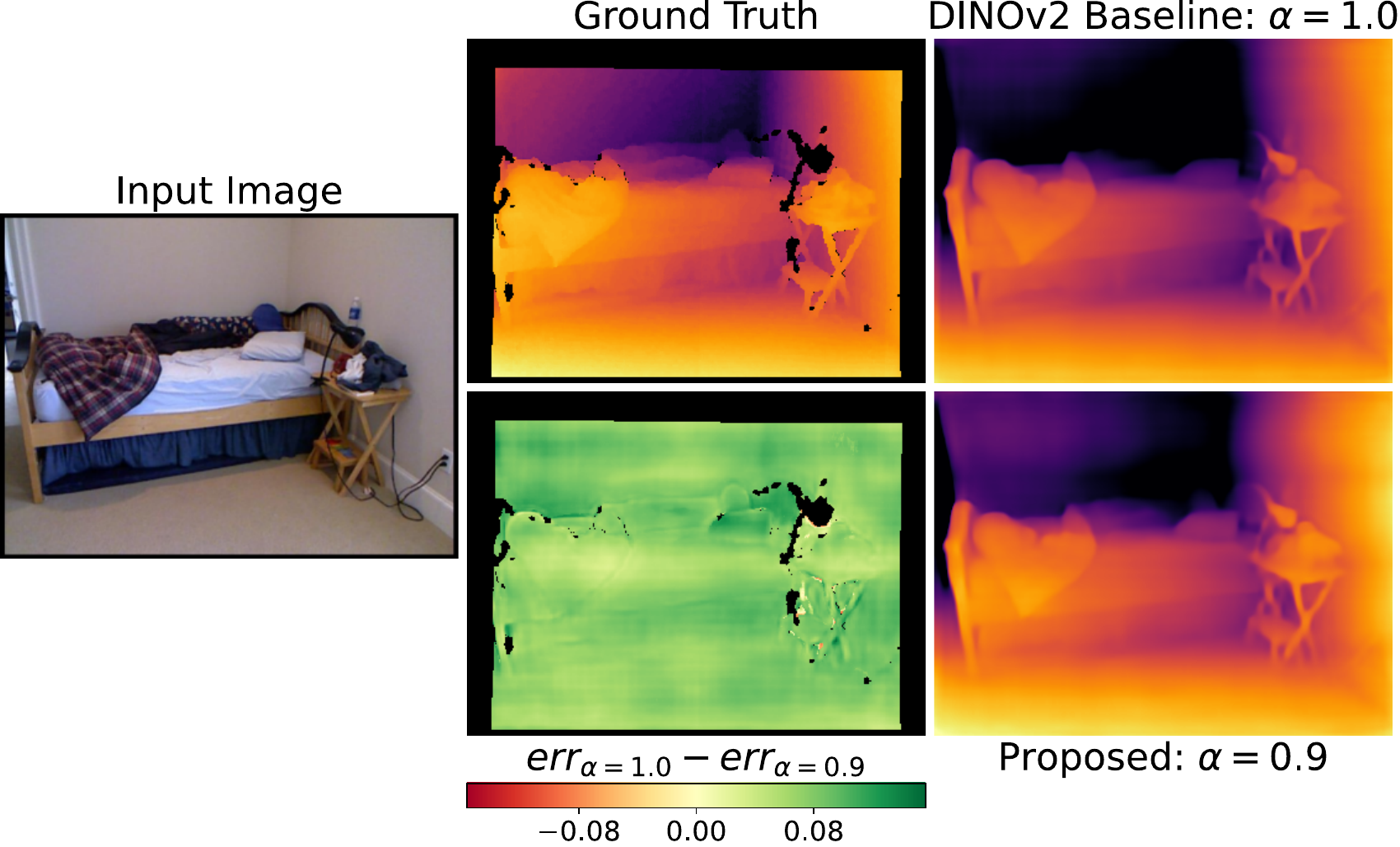}
\resizebox{.7\columnwidth}{!}{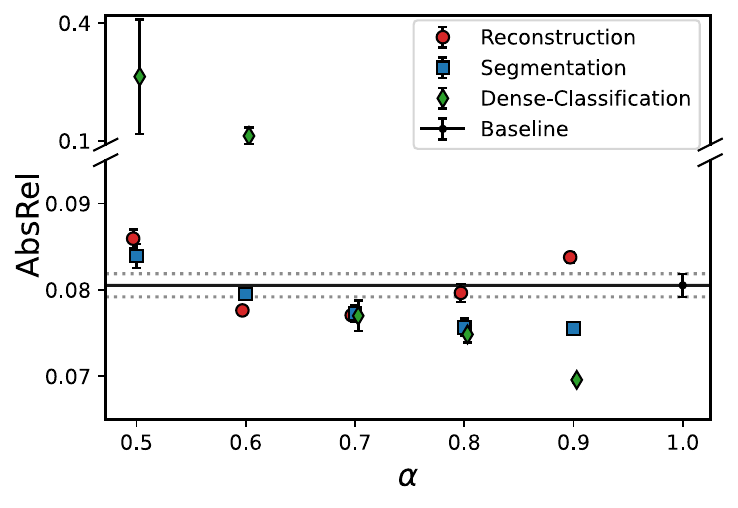}
\caption{(Top) NYUv2 results with MIX6 auxiliary MLDC. From left to right: input, ground truth, error difference (w.r.t. ground truth) between the DINOv2 baseline and ours. Green indicates ours is better, while red vice versa. (Bottom) AbsRel ($\downarrow$) for varying values of the task focusing parameter $\alpha$ with multiple tasks (markers). The solid and dashed lines represent the mean and standard error of DINOv2, respectively.}
\label{fig:eye-catching}
\end{figure}

\section{Introduction}
\label{sec:intro}
Monocular depth estimation (MDE) is a well-established task in computer vision, with possible applications ranging from autonomous vehicles to augmented reality. It is inherently data-hungry, necessitating extensive, high-quality labeled datasets for effective training. Yet, the procurement of such datasets poses a significant challenge, often being a costly and time-consuming endeavor.

Recent advances in monocular depth estimation have been driven by suitable adoption of Vision Transformers (ViT) \cite{vaswani2017attention,dosovitskiy2020image}. In particular DPT \cite{ranftl2021vision} is the first attempt to relative MDE using ViT. Recent architectures, like MiDaS 3.1 \cite{birkl2023midas} and ZoeDepth \cite{bhat2023zoedepth} extend DPT to multiple ViT backbones and metric monocular depth prediction, respectively. Furthermore, DINOv2 \cite{oquab2023dinov2} is a vision foundation model with excellent zero-shot MDE capabilities when combined with a DPT decoder, and Depth Anything \cite{depthanything} fine-tunes this into an MDE foundation model using a large composition of labeled MDE and unlabeled vision data.

Vision transformers have been primarily employed for individual tasks or in self-supervised learning paradigms. Recently, their application in Multi-Task Learning (MTL) is a growing area of research. In particular, there is an increasing interest in using a single model to predict multiple dense tasks, often employing MTL approaches that propose either a unified architecture \cite{kolesnikov2022uvim, yang2023polymax, ning2023all, bhattacharjee2023vision, jha2021s, park2022multi} or methods to balance the loss functions during the training \cite{wang2023images, sener2018multi, kendall2018multi, chen2018gradnorm, guo2018dynamic, liu2019end, bansal2023semantics}. While MTL has been applied to MDE in prior studies \cite{taghavi2024swinmtl, li2024learning, cuevas2024efficient, landgraf2024efficient}, we propose Multi-Label Depth Classification (MLDC) as an auxiliary task to improve MDE.

Motivated by the limited availability of high-quality labeled MDE data, we propose a data-efficient and resource-considerate approach that leverages related vision datasets without requiring extensive fine-tuning of a pre-trained foundation model. This strategy improves MDE performance while reducing the reliance on MDE labels.

Our approach leverages a frozen DINOv2 ViT Giant model \cite{oquab2023dinov2} as a feature extractor, bypassing the need for fine-tuning. We jointly train a shared DPT decoder \cite{ranftl2021vision} with auxiliary datasets from related tasks to improve MDE. We illustrate the qualitative and quantitative improvements of our method over the DINOv2 baseline in Figure \ref{fig:eye-catching}. 

Our key contributions are summarized as follows:
\begin{itemize}[noitemsep]
\item We propose an alternating training scheme leveraging auxiliary non-MDE datasets from related vision tasks to boost the MDE downstream task. This improves the MDE performance by weighting MDE steps more than auxiliary ones through our \textit{task-focusing} parameter $\alpha$;
\item To the best of our knowledge, we are the first to use Multi-Label Dense Classification (MLDC) as an auxiliary task for MDE. 
\item We find that MLDC frequently outperforms semantic segmentation as auxiliary task, suggesting that classification aspects may be more beneficial than spatial details, especially with low-quality segmentation labels.
\item We thoroughly test our method across various MDE datasets, using multiple auxiliary datasets and tasks. Our results show an average quality gain of 11\% compared to the DINOv2 baseline on in-domain datasets, confirming the robustness of our method;
\item We show that our method enhances the data efficiency of DINOv2, allowing for a reduction in MDE training data of 80-99\%, while still improving over DINOv2;
\end{itemize}

These contributions represent a novel advancement in the field, both algorithmically and scientifically.
    
Notably, compared to the recent Depth Anything \cite{depthanything}, which reports no improvements when jointly training their model (using DINOv2 ViT-L encoder) together with semantic segmentation data using task-specific DPT decoders, our method successfully leverages auxiliary tasks to enhance MDE. In particular, our approach keeps the larger DINOv2 ViT-G model frozen and jointly trains a single shared DPT decoder with an auxiliary task to improve MDE. 
In addition, our method has a reduced training cost by design as it does not fine-tune the backbone, while being bounded by its quality.
In this sense, we see Depth Anything and methods that fine-tune the backbone as orthogonal to our work. 

\section{Related Work}
\label{sec:related_work}

\mysubsubsection{Monocular Depth Estimation.} 
Recent works on Monocular Depth Estimation \cite{ranftl2020towards, bhat2023zoedepth, lee2019big, lee2019monocular, lizhenyu2022binsformer, lizhenyu2022depthformer, liu2015deep, nekrasov2019real, oquab2023dinov2, poggi2020uncertainty, shelhamer2016fully, xu2018structured, wang2020sdcdepth} primarily use Vision Transformers \cite{vaswani2017attention,dosovitskiy2020image}. In particular DPT \cite{ranftl2021vision}, also known as MiDaS 3.0, is the first attempt to relative monocular depth estimation using Vision Transformers, adapting the original MiDaS CNN-architecture \cite{ranftl2020towards}. DPT has been used as a baseline for recent MDE SOTA architectures, for instance, MiDaS 3.1 \cite{birkl2023midas} which shows performance of multiple ViT backbones in the MiDaS model, and ZoeDepth\cite{bhat2023zoedepth}, which combines DPT for relative depth estimation with a new module for metric depth prediction. Furthermore, DINOv2 \cite{oquab2023dinov2} proposes a general-purpose vision foundation model that can be used together with DPT decoder to build a powerful zero-shot MDE predictor. Lastly, the recent Depth Anything \cite{depthanything} builds on top of DINOv2 and trains using a large combination of labeled MDE and unlabeled vision datasets. These works are task-specific and therefore miss the potential synergies between different vision tasks. In this paper, we show that the representations extracted by DINOv2 can be leveraged as a strong foundation for multi-task dense prediction, without further fine-tuning DINOv2.

\mysubsubsection{Multi-Task Dense Prediction.} Multi-Task Learning (MTL) \cite{caruana1997multitask,vandenhende2021multi,zhang2021survey,ruder2017overview} leverages the idea that related tasks can provide complementary insights and improve the overall model representations during training, without the need for training multiple separate models. Multi-task dense prediction pertains to the domain of MTL concentrated on dense vision tasks. Through this paper, the abbreviation MTL is employed to denote methods related to Multi-Task Dense Prediction. Research in MTL can be broadly classified into two primary categories. The first \cite{yang2023polymax,wang2023images,bhattacharjee2023vision,kolesnikov2022uvim,ning2023all,li2022learning} focuses on refining network architectures to optimize information sharing across tasks and to enhance task-specific representations. For instance, PolyMax \cite{yang2023polymax} handles diverse prediction tasks effectively, showcasing adaptability and spatial data handling. Painter \cite{wang2023images} leverages in-context learning with an innovative image-centric method. 
The second category \cite{sener2018multi,kendall2018multi,chen2018gradnorm,liu2019end,guo2018dynamic,jha2021s,bansal2023semantics} involves task-balancing optimization, where tasks synergistically learn a single model for solving multiple tasks. Sener et al.'s framework \cite{sener2018multi} fine-tunes task equilibrium, promoting synchronized learning. Kendall et al. \cite{kendall2018multi} address task uncertainty via dynamic weighting mechanism based on confidence levels. Chen et al. \cite{chen2018gradnorm} explore gradient contributions for adaptive loss balancing. Liu et al. \cite{liu2019end} advocate an end-to-end approach, emphasizing integrated learning workflows. 
Guo et al. \cite{guo2018dynamic} propose dynamic task prioritization. 

The methods discussed so far aim to learn multiple tasks simultaneously, whereas we propose an auxiliary learning approach  \cite{Dery2022AANGAA,pmlr-v162-chen22y,Rottmann21,Liu_2023_WACV} where auxiliary tasks are used only in the learning phase to improve MDE performance and are discarded afterwards. This approach allows us to show the positive impact of auxiliary tasks on improving MDE without direct reliance on MTL comparisons.

\mysubsubsection{Cross-Task Relations.} The investigation into cross-task relations \cite{wang2019neural, zamir2018taskonomy, saha2021learning, fifty2021efficiently, wang2021domain, casser2019depth, standley2020, achille2019task2vec, pal2019zero, dwivedi2019representation} goes beyond understanding individual tasks in isolation, aiming to uncover how tasks mutually inform, and enhance each others. Zamir et al.'s Taskonomy \cite{zamir2018taskonomy}, provides a systematic framework that maps relationships between different vision tasks. Fifty et al. introduced Task Affinity Grouping \cite{fifty2021efficiently}, which clusters related tasks to maximize learning efficiency and performance, providing a pragmatic approach to task integration. Wang et al. \cite{wang2021domain} emphasize how cross-task relations can be domain-specific and used for improved model performance. Saha et al. \cite{saha2021learning} present methodologies for models to effectively learn from multiple task interactions. 

In contrast to these works, our focus is not on discovering general cross-task relations, but to leverage data from specifically chosen vision tasks, known to be complementary to MDE, to improve the MDE quality.

\section{Proposed Method}
\label{sec:method}

Monocular Depth Estimation (MDE) is frequently impeded by limited availability of high-quality labeled datasets. Images in publicly available datasets often contain many invalid regions (represented in black in images throughout the paper), where the correct label is undefined or out of range. Additionally, datasets may come with their own pre-processing procedure, where, depending on the context, the distance is limited to specific ranges. MDE methods address this issue by mixing MDE datasets and scaling their size, while MTL methods attempt to train a single model to predict multiple tasks, improving on some or all of them. These methods may suffer from the "task interference" problem, where gradients from different tasks conflict and hinder each other's progress \cite{caruana1997multitask}.

\begin{figure*}[tb]
\centering
\includegraphics[width=0.8\textwidth]{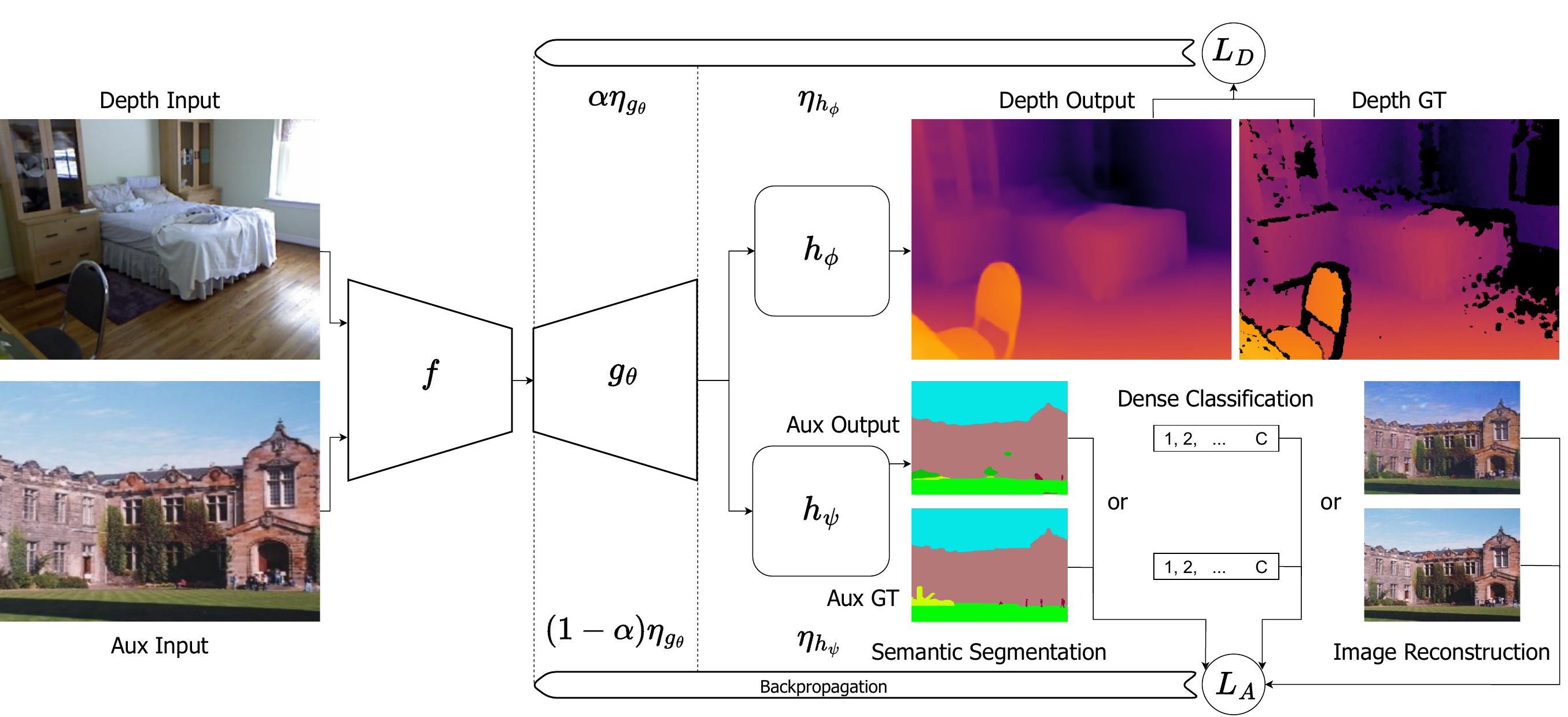}
\caption{Overview of the proposed training pipeline. We use a frozen pre-trained DINOv2 ViT-G backbone ($f$) as a feature extractor and jointly train a DPT decoder ($g_\theta$) with auxiliary datasets from related vision tasks (semantic segmentation, dense classification, or image reconstruction). We train 2 task-specific heads: MDE ($h_\phi$) and auxiliary ($h_\psi$). We alternate MDE steps and auxiliary steps. During backpropagation, each head has its own learning rate ($\eta_{h_\phi}$ and $\eta_{h_\psi}$), while the decoder shares a common learning rate $\eta_{g_\theta}$, scaled by $\alpha$ for MDE and $1 - \alpha$ for the auxiliary task.}
\label{fig:method-diagram}
\end{figure*}

Motivated by these limitations, we propose a method that bridges the gap between conventional MDE techniques and the prevalent methodologies within the MTL domain. Differently from these works, we use datasets from related vision tasks to boost the performance of a frozen pre-trained foundation model on the MDE downstream task, disregarding the auxiliary outcomes. We select DINOv2 ViT-Giant \cite{oquab2023dinov2} as backbone due our requirement for a high-quality and robust feature extractor. We jointly train a shared DPT \cite{ranftl2021vision} decoder by alternating MDE steps and auxiliary task steps, while prioritizing MDE. Differently from Depth Anything \cite{depthanything}, we share a single DPT decoder, and only separate the smaller task-specific heads. This allows us to have an additional shared model component and to freeze the larger pre-trained backbone, while still benefiting from joint training with auxiliary datasets and tasks. Our training method is summarized in Figure~\ref{fig:method-diagram}.

Formally, let $D = {(x_i, y_i)}^N_{i=1}$ be an MDE training dataset with $N$ samples, $A = {(x_k, y_k)}^M_{k=1}$ be an auxiliary training dataset from a related task with $M$ samples, $V = {(x_l, y_l)}^L_{l=1}$ an MDE test dataset with $L$ samples, where $x_.$ are input images and $y_.$ the respective task ground truths. Further, let $f$ be a frozen pre-trained foundation model, $g_\theta$ a decoder with shared parameters $\theta$ and $h_\phi$ and $h_\psi$ respectively the depth and auxiliary task heads with parameters $\phi$ and $\psi$, our goal is to improve the decoder quality on the main downstream task (MDE) by jointly training it with external auxiliary datasets from a related vision task. 
More specifically, let $d_{.}(x) = h_{\phi}(g_{\theta}(f(x))$ be the depth predictor with parameter states $\theta$ and $\phi$, let $\mathcal{M}(V, d_{.})$ be a set of MDE validation metrics to minimize on all test samples $(x_l, y_l) \in V$ using the depth predictor $d_{.}$. Then, let 
\begin{equation}
    m_b = \min \mathcal{M}(V, d_b)
\label{eq:problem}
\end{equation}
be the smallest test metrics achieved by the DINOv2 baseline $d_b$, obtained with the following minimization problem:
\begin{equation}
    d_b = \argmin_{\theta, \phi} \mathcal{L}_D (\hat{y}_i, y_i)
\label{eq:single_training}
\end{equation}
where $\hat{y}_i = d_{.}(x_i)$ is a prediction on a depth training sample $(x_i, y_i) \in D$ with parameters $\theta$ and $\phi$ updated during training, and $\mathcal{L}_D$ is an MDE training loss.

Let $a_j = h_{\psi}(g_{\theta}(f(x))$ be the predictor on an auxiliary related vision task with parameter states $\theta$ and $\psi$, we look for model $d_j$ that best performs on the MDE test dataset $V$ with a set of metrics $\mathcal{M}$, while jointly trained on an auxiliary training dataset $A$ from a related task
\begin{gather}
    d_j = \argmin_{\theta, \phi, \psi} (\mathcal{L}_D (\hat{y}_i, y_i) , \mathcal{L}_A (\hat{y}_k, y_k)) \\
    \text{s.t. }\; \mathcal{M}(V, d_j) \leq m_b \nonumber
\label{eq:joint_training}
\end{gather}
where $\hat{y}_k = a_j(x_k)$ is a prediction on an auxiliary training sample $(x_k, y_k) \in A$ with parameters $\theta$ and $\psi$ updated during training, and $\mathcal{L}_A$ is an auxiliary task training loss.

In order to train (\ref{eq:joint_training}), we define a joint global gradient step as two consecutive task-specific gradient steps. In particular, we first apply a depth gradient step
\begin{equation}
    \theta_{t_D} = \theta_{t-1} - \alpha \eta_{g_\theta} \frac{1}{B_D}\sum^{B_D}_{i = 1} \nabla_{\theta_{t-1}, \phi_{t-1}} \mathcal{L}_D (\hat{y}_i, y_i)
\label{eq:gradstep_depth}
\end{equation}
and then an auxiliary task gradient step
\begin{equation}
    \theta_{t} = \theta_{t_D} - (1-\alpha) \eta_{g_\theta} \frac{1}{B_A}\sum^{B_A}_{k = 1} \nabla_{\theta_{t_D}, \psi_{t-1}} \mathcal{L}_A (\hat{y}_k, y_k) 
\label{eq:gradstep_aux}
\end{equation}
where $t-1$, $t_D$ and $t$ respectively represent the latest global training step, the intermediate MDE step and the updated global step, $B_D$ and $B_A$ are respectively a batch of depth samples $(x_i, y_i) \in D$ and auxiliary task samples $(x_k, y_k) \in A$, $\eta_{g_\theta}$ is the overall decoder learning rate for a full gradient step, weighted by the \textit{task-focusing} parameter $\alpha$, deciding how much of the overall decoder gradient step focuses on the main and auxiliary tasks. This ensures that the total learning rate for the joint scheme is comparable to that of the baseline, which we tested to be optimal (see \Cref{sec:lr_tuning}).
When $\alpha=1$, our training method is reduced to baseline MDE, and when $\alpha=0$, it becomes solely an auxiliary task. Lastly, note that (\ref{eq:gradstep_depth}) and (\ref{eq:gradstep_aux}) omit the gradient step for the task-specific heads $h_{\phi}$ and $h_{\psi}$, which use their own learning rates, $\eta_{\phi}$ and $\eta_{\psi}$ respectively (here $\eta_{\phi} = \eta_{\psi}$).

\mysubsubsection{Auxiliary tasks.} 
We consider three tasks in our evaluations. (1) Semantic segmentation, for which we use a CrossEntropy (CE) loss. (2) \textit{Multi-Label Dense Classification} (\textit{MLDC}), defined as the classification task that can be computed out of semantic segmentation outputs $y_s$ and labels $\hat{y}_s$. We average along the spatial dimensions of $y_s$, obtaining a multi-label vector $y_c = \frac{1}{HW}\sum_{i=1}^{H}\sum_{j=1}^{W}y_{s_{i,j}}$, taking the unique classes $\hat{y}_c$ of $\hat{y}_s$ and computing BinaryCrossEntropy (BCE) loss between $y_c$ and $\hat{y}_c$. We include this task to disentangle the contributions of the classification and positioning parts of the semantic segmentation task. (3) Lastly, we use image reconstruction by using 3 output channels (RGB) and computing the MeanSquaredError (MSE) loss w.r.t. the original input images. We include this task to understand whether task labels are needed at all.

\section{Experimental Results}
\label{sec:experimental_results}

This section presents a comprehensive evaluation of our method. We begin by describing the training details, the studied auxiliary tasks, MDE and auxiliary datasets, and their pre-processing steps. We then evaluate the efficacy of our method, investigating three main areas: the impact of different auxiliary datasets in improving MDE quality; the effectiveness of jointly training across a variety of MDE datasets; and the task relevance, exploring the impact of the auxiliary task selection for the improvement of MDE.

\mysubsubsection{Training details.} 
We adopt the training procedure of DINOv2 \cite{oquab2023dinov2}, both for our main task (MDE) and for our auxiliary task. In practice, we duplicate the optimizer and learning rate scheduler, and then we scale the MDE DPT learning rate by $\alpha$ and the auxiliary task learning rate by $1 - \alpha$, as depicted in \Cref{fig:method-diagram}. 
In \Cref{fig:eye-catching} (right) we study the impact of the hyperparamether $\alpha$ on MIX6, a dataset we introduce below, see Table~\ref{tab:datasets_aux}. Notably, the figure reveals that values greater than $0.6$ are good candidates for achieving improvements on MDE. Throughout the experimental section, we adopt $0.9$ as it showed the best performance. However, this may vary depending on the dataset, task and training procedure. The general optimal selection of $\alpha$ is complementary and not the main goal of this work. Lastly, note that for all experiments we report mean and standard error of 4 independent runs (with different random but fixed seeds). Further training details, and hardware details are provided in Appendices \ref{app:training_details} and \ref{app:hardware_details}, respectively.

\mysubsubsection{Auxiliary Datasets.} 
We consider the auxiliary datasets reported in \Cref{tab:datasets_aux}. We use the pre-processing steps in the \textit{mmsegmentation} framework \cite{mmseg2020}. Additionally, we compose a dataset out of these, which we name \textit{MIX6}. This dataset is used in combination with dataset-specific prediction heads, to avoid the need to join the classes into a unique class space. Each composing dataset keeps its original pre-processing steps. We excluded Taskonomy from MIX6 as its relatively poor segmentation label quality might negatively impact the MDE performance \cite{yang2023polymax}. 

\begin{table}[tb]
\centering
\caption{Adopted auxiliary datasets. We name their union MIX6.}
\setlength{\tabcolsep}{3pt}
\begin{tabular}{lllll}
\toprule
Name       & Scene Type   & Classes & \makecell[l]{Train \\ Size}    \\ \hline
ADE20K \cite{Zhou2017ADE20K} &
  Indoor, Outdoor &
  150 &
  20K \\ 
  
SUN RGBD \cite{song2015sun} &
  Indoor &
  37 &
  5.2K \\ 
  
Cityscapes \cite{Cordts2016Cityscapes}
    & Driving
    & 34
    & 2.9K \\ 

COCO-Stuff \cite{caesar2018cvpr} &
  Indoor, Outdoor &
  171 &
  118K \\ 

Pascal VOC \cite{Everingham10} &
  Indoor, Outdoor &
  21 &
  8.5K \\ 

Pascal Context \cite{mottaghi_cvpr14} &
  Indoor, Outdoor &
  59 &
  5K \\    
\bottomrule
\end{tabular}%
\label{tab:datasets_aux}
\end{table}

\mysubsubsection{MDE datasets.}
\label{sec:mde_datasets}
We use the MDE datasets reported in \Cref{tab:datasets_depth}. In particular, for NYUv2 and SUN RGBD we use the pre-processing steps adopted by DINOv2 \cite{oquab2023dinov2}, while for the other indoor/outdoor datasets we use the NYUv2 pre-processing, without the NYUCrop. For DIODE Outdoor we limit the depth to 80 meters \cite{depthanything}, while for Matterport3D we first resize the images to match the NYU images size (480x640). For Matterport3D and Taskonomy we report results on the validation set. To maintain a proportional scale between MDE and auxiliary datasets, we randomly select 10\% of the images for Taskonomy. 

For KITTI we use the configurations in the \textit{Monocular Depth Estimation Toolbox} \cite{lidepthtoolbox2022}. 
Note that KITTI evaluates a different case than the remaining datasets: images are taken from autonomous driving and street scenes and therefore are semantically and structurally different\footnote{wider images (352x704), large outdoor scenes, and poor label quality.}. 
Nonetheless, we include it to observe the effects of using mainly out-of-domain auxiliary datasets on MDE performance.

\begin{table}[tb]
\centering
\caption{MDE datasets used in this work, where SL=Structured Light; TOF=Time-of-Flight; SCS=Stereo Camera Sensing. For more information about the sensor names, we refer the reader to the work of Lopes et al.\cite {2022arXiv220105761L}. For MatterPort3D and Taskonomy we test on the validation split.}
\setlength{\tabcolsep}{3pt}
\begin{tabular}{lllllll}
\toprule
Name        & \makecell[c]{Scene \\ Type}      & \makecell[c]{Sensor \\ Type} & \makecell[c]{Train \\ Size} & \makecell[c]{Test \\ Size}     \\ \hline
NYUv2 \cite{silbermanNYU} &
  Indoor &
  SL &
  24.2K &  654 \\ 
  
SUN RGBD \cite{song2015sun} &
  Indoor &
  SL, TOF &
  5.2K &  5K \\ 
  
MatterPort3D \cite{Matterport3D} &
  Indoor &
  SCS, SL &
  144K & 19.2K \\ %
  
Taskonomy \cite{zamir2018taskonomy}  & Indoor          & SL 
& 3.2M & 498K \\ 
DIODE \cite{DIODE}      & \makecell[l]{Indoor \\ Outdoor} & LiDAR                                         
& \makecell[l]{8.5K \\ 16.8K} & \makecell[l]{325 \\ 446} \\ 
KITTI \cite{Geiger2013IJRR}     & Driving         & \makecell[l]{SCS, \\ LiDAR}   
& 23K &  652 \\             
\bottomrule
\end{tabular}%
\label{tab:datasets_depth}
\end{table}

\subsection{Enhancement via Multi-Source Auxiliary Tasks}
We study whether we can compensate a decrease in depth estimation information during training with an increase in information from an auxiliary vision task and an external dataset. In other words, if downscaled MDE steps can be compensated by complementary upscaled auxiliary task steps from an external dataset. We conduct extensive experimental evaluations using the training method described in \Cref{sec:method}. Results show that not only can we compensate MDE information with an auxiliary task, but we also improve the overall MDE quality. This suggests that our method can be used as "augmentation" for the main downstream task, while disregarding the auxiliary task outcomes. In the following paragraphs, we show that our method succeeds independently from the auxiliary dataset, consistently improving MDE performance when using in-domain auxiliary data. Lastly, we report observations on the auxiliary task relevance.

\begin{figure*}[tb]
    \centering
    \resizebox{.85\textwidth}{!}{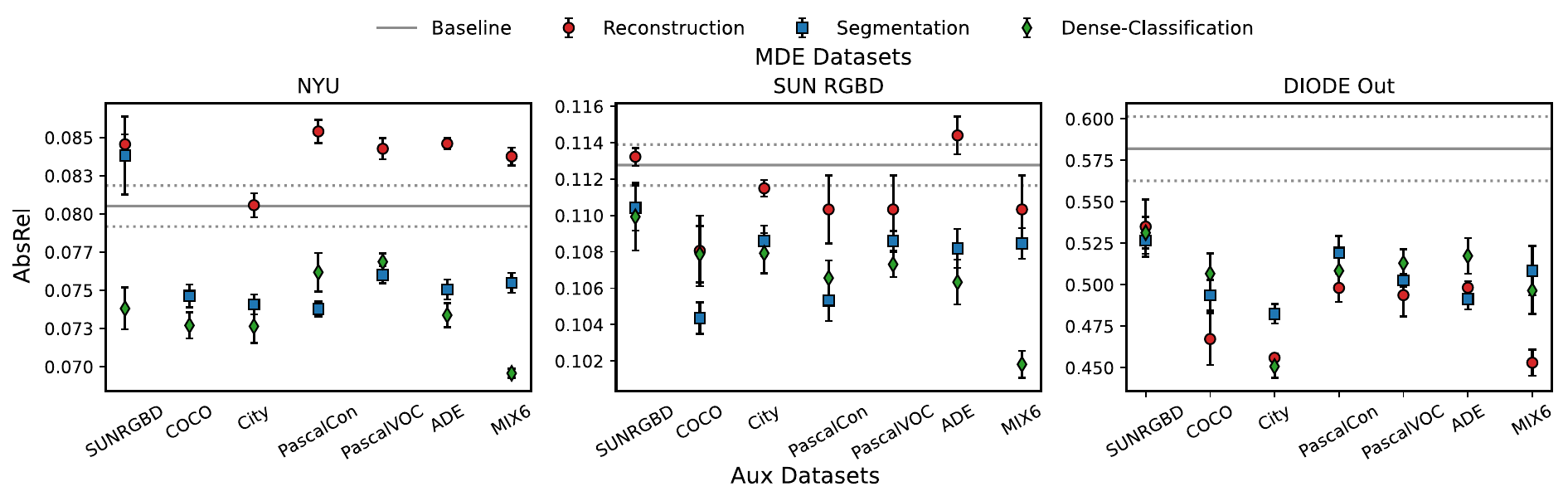}
    \caption{Absolute Relative Error (AbsRel) of MDE on NYUv2, SUN RGBD and DIODE Outdoor using the DINOv2 baseline and our method with multiple auxiliary datasets and tasks. Dots and bars depict the mean and standard error of AbsRel ($\downarrow$), respectively.} 
    \label{fig:nyu_compare_aux_last}
\end{figure*}

\mysubsubsection{Auxiliary dataset choice.}
Figure \ref{fig:nyu_compare_aux_last} illustrates Absolute Relative Error on NYUv2, SUN RGBD and DIODE Outdoor of the baseline and our method with each auxiliary dataset listed in \Cref{tab:datasets_aux}, including their mix, MIX6. The results confirm the enhancement of MDE quality on the three datasets, irrespective of the chosen auxiliary dataset. Notably, MIX6 consistently outperforms others, emerging as the most effective auxiliary dataset for both NYUv2 and SUN RGBD, leading us to use it subsequent experiments unless differently specified. The consistent outperformance of MIX6 suggests that the diverse combination of auxiliary datasets within MIX6 offers a set of cues to enhance MDE quality, with each dataset contributing unique information. We note that for DIODE Outdoor using Cityscapes in dense classification works even a bit better than MIX6.

\mysubsubsection{Results across multiple MDE datasets.}
We extend our experiments to a variety of MDE datasets, covering indoor, outdoor and driving scenes, as reported in \Cref{tab:datasets_depth}. We here use MIX6 as auxiliary dataset as it generally performed the best (\Cref{fig:nyu_compare_aux_last}).
From \Cref{tab:mde_mix6_alpha09}, we see that our method consistently improves MDE quality on indoor and outdoor datasets by an average of $\sim11\%$, indicating that our method is robust to the MDE dataset selection and that MIX6 provides useful information mainly but not exclusively when used as auxiliary training dataset for MLDC. Furthermore, \Cref{tab:mde_mix6_alpha09} demonstrates that incorporating MIX6 as auxiliary dataset for KITTI does not yield beneficial results. We hypothesize that this outcome is due to two main factors: firstly, the inherent challenges presented by KITTI dataset (see \Cref{sec:mde_datasets}), and secondly, the out-of-distribution characteristics of the MIX6 dataset, which is dominated by non-driving scenes. Consequently, we have chosen to exclude KITTI from further evaluations. 

When also considering the out-of-distribution performance on KITTI, our method achieves an average $\sim9\%$ AbsRel gain compared to the DINOv2 baseline. 

\begin{table*}[tb]
\sisetup{
  table-align-uncertainty=true,
  separate-uncertainty=true,
  detect-all = true,
}
\newrobustcmd{\rankfirst}{\fontseries{b}\selectfont}
\newrobustcmd{\ranksecond}{\fontshape{sl}\selectfont}
\centering
\setlength{\tabcolsep}{3pt} 
\caption{AbsRel $\times 10^4$ ($\downarrow$) scores and percentage gain of the best task w.r.t. the DINOv2 baseline on various indoor and outdoor (top 6) and driving scenes (bottom 2) datasets, using different auxiliary tasks. The indoor-outdoor dominated MIX6 dataset with $\alpha=0.9$ is used for all the experiments. The best and second-best results are highlighted in \textbf{bold} and \textit{italic}, respectively.}
    \begin{tabular}{
     @{} l c *{4}{S[table-format=4.0(3),detect-weight,mode=text]} c @{} 
    }
    \toprule
    \multirow{2}{*}{\begin{tabular}[c]{@{}l@{}}MDE\\ Datasets\end{tabular}} & \multirow{2}{*}{\begin{tabular}[c]{@{}c@{}}In\\ MIX6\end{tabular}} & \multicolumn{1}{c}{} & \multicolumn{3}{c}{Aux Tasks} & \multicolumn{1}{c}{} \\ \cmidrule(lr){4-6}
                & & {DINOv2}       & {Classification}  & {Segmentation}   & {Reconstruction} & {Gain \%}\\ \hline
    NYUv2       & \xmark  &    809\pm 10   & \rankfirst  696\pm 3   & \ranksecond  755 \pm 6  & 838\pm 7    & 13.9         \\
    SUN RGBD     & \cmark &  1128 \pm 11   & \rankfirst 1024\pm 10   & \ranksecond 1069 \pm 9  &  1111 \pm 5   & 9.2         \\
    Matterport3D & \xmark & 1874\pm 19   & \rankfirst 1728\pm 9   & \ranksecond 1793 \pm  13  &  1805\pm 6   & 7.8         \\
    Taskonomy   & \xmark  & \ranksecond 1506\pm 13   & \rankfirst 1481\pm 4   &   1567\pm 12  &  1543\pm 6      & 1.7      \\
    DIODE In    & \xmark  &     3588\pm 19   &  \rankfirst  3239\pm26   &   3451\pm 47  &   \ranksecond 3368\pm31    & 9.7        \\
    DIODE Out   & \xmark  &     5820\pm223   &            \ranksecond 4965\pm162   &   5085\pm172  &  \rankfirst   4530\pm91 & 22.2 \\ 
    KITTI       & \xmark  &\rankfirst   605\pm5 &        \ranksecond    609\pm  7   &    646\pm3    &   642\pm6   & -0.6          \\
    \bottomrule
    \end{tabular}%
\label{tab:mde_mix6_alpha09}
\end{table*}

\mysubsubsection{Task relevance.}
We ask the question: "Is the contribution coming from the auxiliary task or simply from adding data?". As depicted in \Cref{fig:nyu_compare_aux_last} and detailed in \Cref{tab:mde_mix6_alpha09}, employing the same dataset with different tasks yields varying results, indicating that improvements in MDE quality do not merely stem from data addition. Remarkably, \Cref{tab:mde_mix6_alpha09} shows the consistent superiority of MLDC as an auxiliary task, outperforming other tasks across datasets, with exceptions for DIODE Outdoor, where image reconstruction is more effective. Interestingly, MLDC often surpasses semantic segmentation, suggesting that identifying classes in an image, without the need for precise object positioning, is sufficient for auxiliary task effectiveness. 
We hypothesize that this holds true, especially when semantic segmentation labels lack positional precision. The broader implication is that semantic segmentation datasets can be repurposed as dense classification datasets to enhance various vision downstream tasks, showcasing the versatility and impact of repurposing datasets for auxiliary tasks. Given these findings, MLDC is selected as the auxiliary task for the subsequent experiments, unless explicitly specified otherwise. We additionally evaluate the method with single-label dense classification as an auxiliary task and show in Appendix \ref{app:single_label} that it also improves MDE quality compared to the baseline, while using multiple labels seems better.

\subsection{Improved Data Efficiency Properties}
\begin{figure*}[tb]
    \centering
    \resizebox{.6\textwidth}{!}{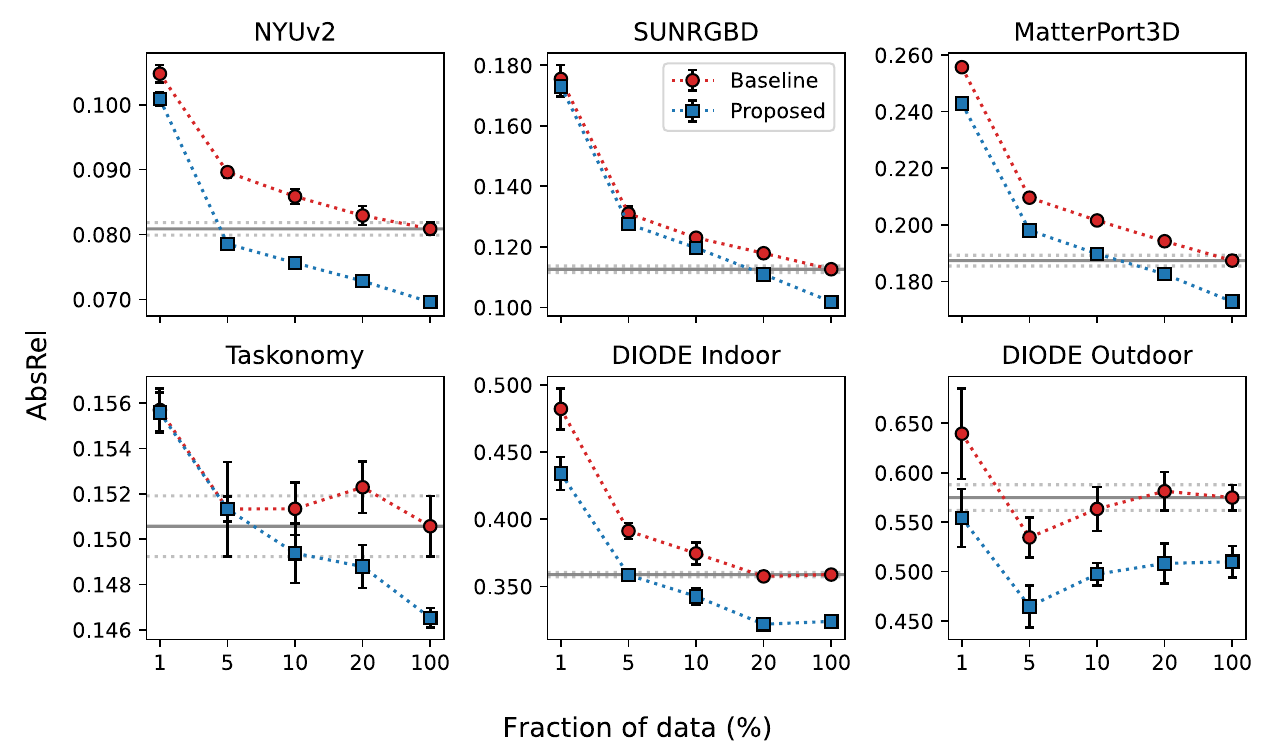}
    \caption{AbsRel of models trained with various fractions of the dataset. The dataset sizes are reported in \Cref{tab:datasets_depth}. DINOv2 baseline (red circles) represents the model trained without auxiliary tasks, whereas Proposed (blue squares) depicts our method jointly trained with MIX6 MLDC auxiliary task with $\alpha=0.9$.} 
    \label{fig:data_efficiency}
\end{figure*}

Given challenges associated with collecting new high-quality labeled MDE datasets, we explore the data efficiency properties of our method by seeking an answer to the following question: "To what extent can we reduce the usage of MDE data while maintaining performance comparable to our full-data baseline?". Our investigations highlight the intrinsic data efficiency of the frozen DINOv2 ViT-G for MDE, up to a certain degree. As shown in \Cref{fig:data_efficiency}, this efficiency is amplified by our method on all tested MDE datasets, allowing from 80\% to 99\% less labeled MDE data usage without decreasing the performance. These promising results suggest that for future MDE dataset collection it may be possible to gather less data, while still being able to compensate for it using auxiliary datasets.

\section{Ablation studies}
We conduct ablation studies on the impact of learning rate scaling, and robustness to backbone configurations.

\subsection{Tuning the Learning Rate with and without Auxiliary Task} 
\label{sec:lr_tuning}
\begin{figure*}[tb]
    \centering
    \resizebox{.65\textwidth}{!}{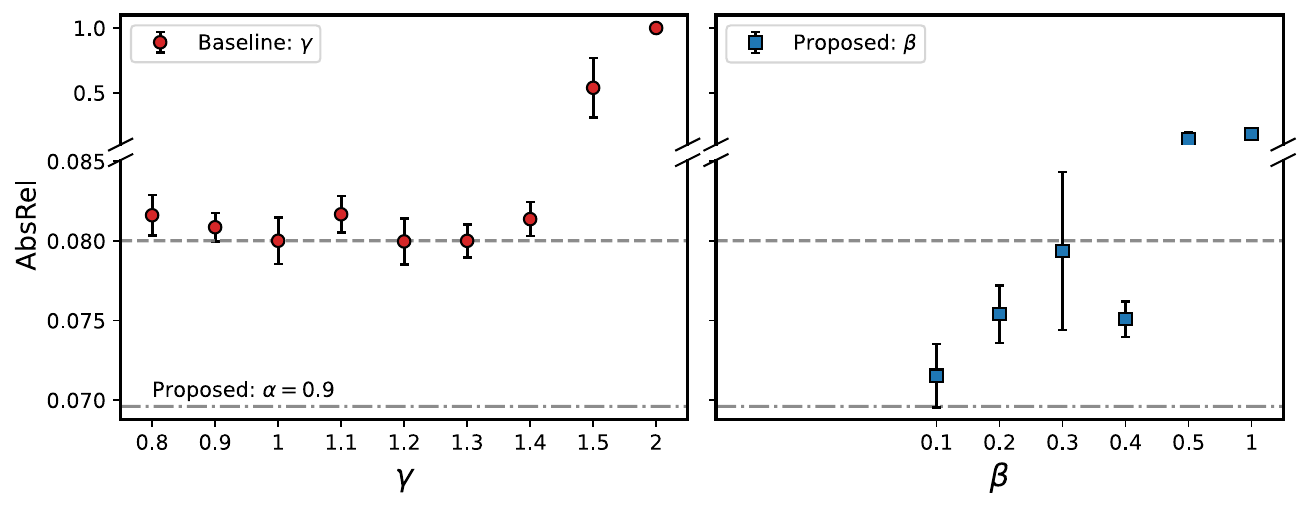}
    \caption{AbsRel ($\downarrow$) when varying the learning rate. (Left) Learning rate tuning for the DINOv2 baseline by a factor $\gamma$. (Right) Our method re-designed to do unscaled depth steps and auxiliary steps scaled by a factor $\beta$, using MIX6 MLDC task. For both plots, dashed and dotted-dashed lines represent the performance of the baseline ($\gamma=1$) and of our method with $\alpha=0.9$, respectively.}
    \label{fig:stl_lr_sensitivity_dam}
\end{figure*}

In order to verify that the contribution comes from the auxiliary task, and not merely from adjusting the learning rate, we repeat the baseline experiments with learning rate scaled by a factor $\gamma$, as reported in \Cref{fig:stl_lr_sensitivity_dam} (Baseline, $\gamma$). The figure shows that the adopted learning rate (scaled by $\gamma=1$) is optimal for MDE on NYUv2, and that differently scaling it produces on par or worse performance.

Furthermore, we motivate our method design by reporting results for an alternative method, where only auxiliary steps are scaled. In other words we re-design our method such that the learning rates for the shared DPT decoder are $\eta_{g_\theta}$ and $\beta\eta_{g_\theta}$, respectively for MDE and the auxiliary task. \Cref{fig:stl_lr_sensitivity_dam} (Proposed, $\beta$) shows that this method can still improve the MDE performance, while being slightly worse when compared to our proposed method, suggesting that using some auxiliary information from related datasets and tasks is beneficial for MDE training, while highlighting the importance of using a small weight for the auxiliary task, to prevent it from overtaking the MDE task.

\subsection{Validation with Depth Anything Backbone} 
We validate our method using the recent Depth Anything model \cite{depthanything} as backbone, which is pre-trained on over 60 million samples, and therefore is a stronger MDE backbone than DINOv2. We follow its original training details, extending the training duration (same as DINOv2 \cite{oquab2023dinov2}) to compensate for the frozen backbone and randomly initialized DPT decoder. \Cref{tab:mde_dam} shows that our method improves when using this backbone, achieving an average 2.5\% quality gain. This suggests that our training scheme can enhance MDE quality across various pre-trained backbones.

\begin{table}[tb]
\sisetup{
  table-align-uncertainty=true,
  separate-uncertainty=true,
  detect-all = true,
}
\newrobustcmd{\rankfirst}{\fontseries{b}\selectfont}
\newrobustcmd{\ranksecond}{\fontshape{sl}\selectfont}
\centering
\setlength{\tabcolsep}{3pt} 
\caption{AbsRel $\times 10^4$ ($\downarrow$) scores on various depth datasets using Depth Anything as baseline and our method with MLDC and $\alpha=0.9$, and percentage gain of our method w.r.t. Depth Anything. The best results are highlighted in \textbf{bold}.}
    \begin{tabular}{l*{0} c*{2}{S[table-format=4.0(3),detect-weight,mode=text]} c @{}}
    \toprule
    {MDE Datasets} & {Depth Anything} & {Ours}  &  {Gain \%} \\ \hline
    NYUv2        & 736\pm15 & \rankfirst 721\pm10 & 2.1 \\
    SUN RGBD     & 1028\pm7 & \rankfirst 1018\pm15 & 1 \\
    Matterport3D & 1885\pm9 & \rankfirst 1843\pm2 & 2.3 \\
    Taskonomy    & 1741\pm15 & \rankfirst 1687\pm13 & 3.1 \\
    DIODE In     & \rankfirst 1993\pm14 & 2073\pm22 & -0.4\\
    DIODE Out    & 3835\pm164 & \rankfirst 3574\pm66 & 6.8 \\ 
    \bottomrule
    \end{tabular}
\label{tab:mde_dam}
\end{table}

\section{Qualitative Results}
\Cref{fig:eye-catching} (top) reports a prediction on a NYUv2 validation sample using the baseline and our method using MIX6 MLDC auxiliary task with $\alpha=0.9$. Note that black regions represent invalid regions, i.e. pixels for which the label is unknown or out of range. We define $err_{m} = | \hat{y}  - y | / \hat{y}$ as the (absolute relative) error map of the prediction $y$ of a method $m$ w.r.t. the ground truth $\hat{y}$. Note that rows in these figures should be considered independently. The figure shows $err_{\alpha = 1.0} - err_{\alpha = 0.9}$, i.e. the difference between the errors of the baseline and our method, respectively, where green depicts areas where our method is closer to the ground truth than the baseline, and red vice versa. In this figure, it is clear that the most distant area (right corner) is lighter and therefore closer to the ground truth. More qualitative results are provided in Appendix \ref{app:qualitative}.
In general, our method's predictions are closer to the ground truth, even though not always these improvements are well visible by human eyesight.

\section{Discussion and Limitations}
Our method presents promising advancements, however it is essential to acknowledge its limitations. 

The selection of auxiliary datasets and tasks plays a pivotal role in improving MDE. We show that MLDC is effective with various datasets. However, the observed task dependency highlights the significance of carefully choosing auxiliary tasks and datasets for optimal outcomes. 

As mentioned in the results section, we report no improvement on KITTI. This can be attributed to several factors. First, as discussed in \Cref{sec:mde_datasets}, driving images are pre-processed differently than other datasets and have poor and sparse labels.
Second, the domain distribution of the KITTI dataset is specific to driving scenes, which is not adequately  represented in our auxiliary datasets. Third, the quantity and variety of auxiliary datasets are crucial for enhancing the model performance. This is evident when comparing our results in \Cref{tab:mde_mix6_alpha09} and the corresponding ablation in \Cref{tab:mde_same_alpha09} in Appendix \ref{app:single_multi}, which shows a decrease in performance across all datasets when using the same dataset exclusively as auxiliary. 
These observations suggest that including of a broader and more diverse range of driving data could enhance the baseline performance for KITTI.

\section{Conclusion}
\label{sec:conclusion}

Monocular Depth Estimation poses challenges  stemming from scarcity of high-quality labeled datasets \cite{depthanything}. We propose joint training of a shared decoder on top of a frozen pre-trained model using auxiliary vision datasets and tasks.

Through our experimental analysis, we show that our method is robust to the auxiliary dataset choice and that we can improve MDE performance on MDE datasets with in-domain auxiliary datasets by an average of 11\% compared to the DINOv2 baseline.Our method demonstrates improved data efficiency for MDE, allowing for at least 80\% MDE data reduction on the tested datasets. This suggests that it may enable quality gains even in scenarios where access to high-quality labeled data is limited. By leveraging auxiliary datasets and a frozen foundation model, our approach improves the quality without the need for extensive re-training. Moreover, our results highlight the influence of semantic segmentation datasets employed as MLDC task and that improvements are due to the selected auxilliary task, not only due to adding more data.

In conclusion, we introduce a compelling strategy for improving MDE by addressing its challenges and demonstrating the potential of using auxiliary datasets. We aim to encourage the exploration of auxiliary task balancing strategies and further studies on auxiliary data and task selection.

\subsection*{Acknowledgements}
Alessio Quercia was funded by the Helmholtz School for Data Science in Life, Earth, and Energy (HDS-LEE). The authors gratefully acknowledge the Gauss Centre for Supercomputing e.V. (\url{www.gauss-centre.eu}) for funding this project by providing computing time through the John von Neumann Institute for Computing (NIC) on the GCS Supercomputer JUWELS \cite{kesselheim2021juwels} at J\"ulich Supercomputing Centre (JSC).

{\small
\bibliographystyle{ieee_fullname}
\bibliography{main}

\begin{thebibliography}{10}\itemsep=-1pt

\bibitem{achille2019task2vec}
Alessandro Achille, Michael Lam, Rahul Tewari, Avinash Ravichandran, Subhransu
  Maji, Charless~C Fowlkes, Stefano Soatto, and Pietro Perona.
\newblock Task2vec: Task embedding for meta-learning.
\newblock In {\em Proceedings of the IEEE/CVF international conference on
  computer vision}, pages 6430--6439, 2019.

\bibitem{EPA_2024}
United States Environmental~Protection Agency.
\newblock Greenhouse gas equivalencies calculator, Mar 2024.

\bibitem{bansal2023semantics}
Nitin Bansal, Pan Ji, Junsong Yuan, and Yi Xu.
\newblock Semantics-depth-symbiosis: Deeply coupled semi-supervised learning of
  semantics and depth.
\newblock In {\em Proceedings of the IEEE/CVF Winter Conference on Applications
  of Computer Vision}, pages 5828--5839, 2023.

\bibitem{bhat2023zoedepth}
Shariq~Farooq Bhat, Reiner Birkl, Diana Wofk, Peter Wonka, and Matthias
  M{\"u}ller.
\newblock Zoedepth: Zero-shot transfer by combining relative and metric depth.
\newblock {\em arXiv preprint arXiv:2302.12288}, 2023.

\bibitem{bhattacharjee2023vision}
Deblina Bhattacharjee, Sabine S{\"u}sstrunk, and Mathieu Salzmann.
\newblock Vision transformer adapters for generalizable multitask learning.
\newblock In {\em Proceedings of the IEEE/CVF International Conference on
  Computer Vision}, pages 19015--19026, 2023.

\bibitem{birkl2023midas}
Reiner Birkl, Diana Wofk, and Matthias M{\"u}ller.
\newblock Midas v3. 1--a model zoo for robust monocular relative depth
  estimation.
\newblock {\em arXiv preprint arXiv:2307.14460}, 2023.

\bibitem{caesar2018cvpr}
Holger Caesar, Jasper Uijlings, and Vittorio Ferrari.
\newblock Coco-stuff: Thing and stuff classes in context.
\newblock In {\em Computer vision and pattern recognition (CVPR), 2018 IEEE
  conference on}. IEEE, 2018.

\bibitem{caruana1997multitask}
Rich Caruana.
\newblock Multitask learning.
\newblock {\em Machine learning}, 28:41--75, 1997.

\bibitem{casser2019depth}
Vincent Casser, Soeren Pirk, Reza Mahjourian, and Anelia Angelova.
\newblock Depth prediction without the sensors: Leveraging structure for
  unsupervised learning from monocular videos.
\newblock In {\em Proceedings of the AAAI Conference on Artificial
  Intelligence}, volume~33, pages 8001--8008, 2019.

\bibitem{Matterport3D}
Angel Chang, Angela Dai, Thomas Funkhouser, Maciej Halber, Matthias Niessner,
  Manolis Savva, Shuran Song, Andy Zeng, and Yinda Zhang.
\newblock Matterport3d: Learning from rgb-d data in indoor environments.
\newblock {\em International Conference on 3D Vision (3DV)}, 2017.

\bibitem{pmlr-v162-chen22y}
Hong Chen, Xin Wang, Chaoyu Guan, Yue Liu, and Wenwu Zhu.
\newblock Auxiliary learning with joint task and data scheduling.
\newblock In Kamalika Chaudhuri, Stefanie Jegelka, Le Song, Csaba Szepesvari,
  Gang Niu, and Sivan Sabato, editors, {\em Proceedings of the 39th
  International Conference on Machine Learning}, volume 162 of {\em Proceedings
  of Machine Learning Research}, pages 3634--3647. PMLR, 17--23 Jul 2022.

\bibitem{chen2018gradnorm}
Zhao Chen, Vijay Badrinarayanan, Chen-Yu Lee, and Andrew Rabinovich.
\newblock Gradnorm: Gradient normalization for adaptive loss balancing in deep
  multitask networks.
\newblock In {\em International conference on machine learning}, pages
  794--803. PMLR, 2018.

\bibitem{mmseg2020}
MMSegmentation Contributors.
\newblock {MMSegmentation}: Openmmlab semantic segmentation toolbox and
  benchmark.
\newblock \url{https://github.com/open-mmlab/mmsegmentation}, 2020.

\bibitem{Cordts2016Cityscapes}
Marius Cordts, Mohamed Omran, Sebastian Ramos, Timo Rehfeld, Markus Enzweiler,
  Rodrigo Benenson, Uwe Franke, Stefan Roth, and Bernt Schiele.
\newblock The cityscapes dataset for semantic urban scene understanding.
\newblock In {\em Proc. of the IEEE Conference on Computer Vision and Pattern
  Recognition (CVPR)}, 2016.

\bibitem{cuevas2024efficient}
Hanz Cuevas-Velasquez, Alejandro Gal{\'a}n-Cuenca, Robert~B Fisher, and
  Antonio~Javier Gallego.
\newblock Efficient multi-task progressive learning for semantic segmentation
  and disparity estimation.
\newblock {\em Pattern Recognition}, 154:110601, 2024.

\bibitem{deng2009imagenet}
Jia Deng, Wei Dong, Richard Socher, Li-Jia Li, Kai Li, and Li Fei-Fei.
\newblock Imagenet: A large-scale hierarchical image database.
\newblock In {\em 2009 IEEE conference on computer vision and pattern
  recognition}, pages 248--255. Ieee, 2009.

\bibitem{Dery2022AANGAA}
Lucio~M. Dery, Paul Michel, Mikhail Khodak, Graham Neubig, and Ameet Talwalkar.
\newblock Aang: Automating auxiliary learning.
\newblock {\em ArXiv}, abs/2205.14082, 2022.

\bibitem{dosovitskiy2020image}
Alexey Dosovitskiy, Lucas Beyer, Alexander Kolesnikov, Dirk Weissenborn,
  Xiaohua Zhai, Thomas Unterthiner, Mostafa Dehghani, Matthias Minderer, Georg
  Heigold, Sylvain Gelly, et~al.
\newblock An image is worth 16x16 words: Transformers for image recognition at
  scale.
\newblock {\em arXiv preprint arXiv:2010.11929}, 2020.

\bibitem{dwivedi2019representation}
Kshitij Dwivedi and Gemma Roig.
\newblock Representation similarity analysis for efficient task taxonomy \&
  transfer learning.
\newblock In {\em Proceedings of the IEEE/CVF Conference on Computer Vision and
  Pattern Recognition}, pages 12387--12396, 2019.

\bibitem{Everingham10}
M. Everingham, L. Van~Gool, C.~K.~I. Williams, J. Winn, and A. Zisserman.
\newblock The pascal visual object classes (voc) challenge.
\newblock {\em International Journal of Computer Vision}, 88(2):303--338, June
  2010.

\bibitem{fifty2021efficiently}
Chris Fifty, Ehsan Amid, Zhe Zhao, Tianhe Yu, Rohan Anil, and Chelsea Finn.
\newblock Efficiently identifying task groupings for multi-task learning.
\newblock {\em Advances in Neural Information Processing Systems},
  34:27503--27516, 2021.

\bibitem{Geiger2013IJRR}
Andreas Geiger, Philip Lenz, Christoph Stiller, and Raquel Urtasun.
\newblock Vision meets robotics: The kitti dataset.
\newblock {\em International Journal of Robotics Research (IJRR)}, 2013.

\bibitem{guo2018dynamic}
Michelle Guo, Albert Haque, De-An Huang, Serena Yeung, and Li Fei-Fei.
\newblock Dynamic task prioritization for multitask learning.
\newblock In {\em Proceedings of the European conference on computer vision
  (ECCV)}, pages 270--287, 2018.

\bibitem{jha2021s}
Ankit Jha, Biplab Banerjee, and Subhasis Chaudhuri.
\newblock S 3 dmt-net: improving soft sharing based multi-task cnn using
  task-specific distillation and cross-task interactions.
\newblock In {\em Proceedings of the Twelfth Indian Conference on Computer
  Vision, Graphics and Image Processing}, pages 1--9, 2021.

\bibitem{kendall2018multi}
Alex Kendall, Yarin Gal, and Roberto Cipolla.
\newblock Multi-task learning using uncertainty to weigh losses for scene
  geometry and semantics.
\newblock In {\em Proceedings of the IEEE conference on computer vision and
  pattern recognition}, pages 7482--7491, 2018.

\bibitem{kesselheim2021juwels}
Stefan Kesselheim, Andreas Herten, Kai Krajsek, Jan Ebert, Jenia Jitsev, Mehdi
  Cherti, Michael Langguth, Bing Gong, Scarlet Stadtler, Amirpasha Mozaffari,
  et~al.
\newblock Juwels booster--a supercomputer for large-scale ai research.
\newblock In {\em International Conference on High Performance Computing},
  pages 453--468. Springer, 2021.

\bibitem{kolesnikov2022uvim}
Alexander Kolesnikov, Andr{\'e} Susano~Pinto, Lucas Beyer, Xiaohua Zhai,
  Jeremiah Harmsen, and Neil Houlsby.
\newblock Uvim: A unified modeling approach for vision with learned guiding
  codes.
\newblock {\em Advances in Neural Information Processing Systems},
  35:26295--26308, 2022.

\bibitem{lacoste2019quantifying}
Alexandre Lacoste, Alexandra Luccioni, Victor Schmidt, and Thomas Dandres.
\newblock Quantifying the carbon emissions of machine learning.
\newblock {\em arXiv preprint arXiv:1910.09700}, 2019.

\bibitem{landgraf2024efficient}
Steven Landgraf, Markus Hillemann, Theodor Kapler, and Markus Ulrich.
\newblock Efficient multi-task uncertainties for joint semantic segmentation
  and monocular depth estimation.
\newblock {\em arXiv preprint arXiv:2402.10580}, 2024.

\bibitem{lee2019big}
Jin~Han Lee, Myung-Kyu Han, Dong~Wook Ko, and Il~Hong Suh.
\newblock From big to small: Multi-scale local planar guidance for monocular
  depth estimation.
\newblock {\em arXiv preprint arXiv:1907.10326}, 2019.

\bibitem{lee2019monocular}
Jae-Han Lee and Chang-Su Kim.
\newblock Monocular depth estimation using relative depth maps.
\newblock In {\em Proceedings of the IEEE conference on computer vision and
  pattern recognition}, pages 9167--9176, 2019.

\bibitem{li2024learning}
Kunqian Li, Xiya Wang, Wenjie Liu, Qi Qi, Guojia Hou, Zhiguo Zhang, and Kun
  Sun.
\newblock Learning scribbles for dense depth: Weakly-supervised single
  underwater image depth estimation boosted by multi-task learning.
\newblock {\em IEEE Transactions on Geoscience and Remote Sensing}, 2024.

\bibitem{li2022learning}
Wei-Hong Li, Xialei Liu, and Hakan Bilen.
\newblock Learning multiple dense prediction tasks from partially annotated
  data.
\newblock In {\em Proceedings of the IEEE/CVF Conference on Computer Vision and
  Pattern Recognition}, pages 18879--18889, 2022.

\bibitem{lidepthtoolbox2022}
Zhenyu Li.
\newblock Monocular depth estimation toolbox.
\newblock \url{https://github.com/zhyever/Monocular-Depth-Estimation-Toolbox},
  2022.

\bibitem{lizhenyu2022depthformer}
Zhenyu Li, Zehui Chen, Xianming Liu, and Junjun Jiang.
\newblock Depthformer: Exploiting long-range correlation and local information
  for accurate monocular depth estimation.
\newblock {\em arXiv preprint arXiv:2203.14211}, 2022.

\bibitem{lizhenyu2022binsformer}
Zhenyu Li, Xuyang Wang, Xianming Liu, and Junjun Jiang.
\newblock Binsformer: Revisiting adaptive bins for monocular depth estimation.
\newblock {\em arXiv preprint arXiv:2204.00987}, 2022.

\bibitem{liu2015deep}
Fayao Liu, Chunhua Shen, and Guosheng Lin.
\newblock Deep convolutional neural fields for depth estimation from a single
  image.
\newblock In {\em Proceedings of the IEEE International Conference on Computer
  Vision}, pages 5162--5170, 2015.

\bibitem{Liu_2023_WACV}
Huan Liu, Zhixiang Chi, Yuanhao Yu, Yang Wang, Jun Chen, and Jin Tang.
\newblock Meta-auxiliary learning for future depth prediction in videos.
\newblock In {\em Proceedings of the IEEE/CVF Winter Conference on Applications
  of Computer Vision (WACV)}, pages 5756--5765, January 2023.

\bibitem{liu2019end}
Shikun Liu, Edward Johns, and Andrew~J Davison.
\newblock End-to-end multi-task learning with attention.
\newblock In {\em Proceedings of the IEEE/CVF conference on computer vision and
  pattern recognition}, pages 1871--1880, 2019.

\bibitem{2022arXiv220105761L}
Alexandre {Lopes}, Roberto {Souza}, and Helio {Pedrini}.
\newblock {A Survey on RGB-D Datasets}.
\newblock {\em arXiv e-prints}, page arXiv:2201.05761, Jan. 2022.

\bibitem{mottaghi_cvpr14}
Roozbeh Mottaghi, Xianjie Chen, Xiaobai Liu, Nam-Gyu Cho, Seong-Whan Lee, Sanja
  Fidler, Raquel Urtasun, and Alan Yuille.
\newblock The role of context for object detection and semantic segmentation in
  the wild.
\newblock In {\em IEEE Conference on Computer Vision and Pattern Recognition
  (CVPR)}, 2014.

\bibitem{silbermanNYU}
Pushmeet~Kohli Nathan~Silberman, Derek~Hoiem and Rob Fergus.
\newblock Indoor segmentation and support inference from rgbd images.
\newblock In {\em ECCV}, 2012.

\bibitem{nekrasov2019real}
Vladimir Nekrasov, Thanuja Dharmasiri, Andrew Spek, Tom Drummond, Chunhua Shen,
  and Ian Reid.
\newblock Real-time joint semantic segmentation and depth estimation using
  asymmetric annotations.
\newblock In {\em 2019 International Conference on Robotics and Automation
  (ICRA)}, pages 7101--7107. IEEE, 2019.

\bibitem{matterportAgreement}
Matthias Niessner.
\newblock Matterport3d eula for academic use, Mar 2024.

\bibitem{ning2023all}
Jia Ning, Chen Li, Zheng Zhang, Chunyu Wang, Zigang Geng, Qi Dai, Kun He, and
  Han Hu.
\newblock All in tokens: Unifying output space of visual tasks via soft token.
\newblock In {\em Proceedings of the IEEE/CVF International Conference on
  Computer Vision}, pages 19900--19910, 2023.

\bibitem{oquab2023dinov2}
Maxime Oquab, Timoth{\'e}e Darcet, Th{\'e}o Moutakanni, Huy Vo, Marc
  Szafraniec, Vasil Khalidov, Pierre Fernandez, Daniel Haziza, Francisco Massa,
  Alaaeldin El-Nouby, et~al.
\newblock Dinov2: Learning robust visual features without supervision.
\newblock {\em arXiv preprint arXiv:2304.07193}, 2023.

\bibitem{pal2019zero}
Arghya Pal and Vineeth~N Balasubramanian.
\newblock Zero-shot task transfer.
\newblock In {\em Proceedings of the IEEE/CVF Conference on Computer Vision and
  Pattern Recognition}, pages 2189--2198, 2019.

\bibitem{park2022multi}
Sangjoon Park and Jong~Chul Ye.
\newblock Multi-task distributed learning using vision transformer with random
  patch permutation.
\newblock {\em IEEE Transactions on Medical Imaging}, 2022.

\bibitem{patterson2021carbon}
David Patterson, Joseph Gonzalez, Quoc Le, Chen Liang, Lluis-Miquel Munguia,
  Daniel Rothchild, David So, Maud Texier, and Jeff Dean.
\newblock Carbon emissions and large neural network training.
\newblock {\em arXiv preprint arXiv:2104.10350}, 2021.

\bibitem{poggi2020uncertainty}
Matteo Poggi, Filippo Aleotti, Fabio Tosi, and Stefano Mattoccia.
\newblock On the uncertainty of self-supervised monocular depth estimation.
\newblock In {\em Proceedings of the IEEE/CVF Conference on Computer Vision and
  Pattern Recognition}, pages 3227--3237, 2020.

\bibitem{ranftl2021vision}
Ren{\'e} Ranftl, Alexey Bochkovskiy, and Vladlen Koltun.
\newblock Vision transformers for dense prediction.
\newblock In {\em Proceedings of the IEEE/CVF international conference on
  computer vision}, pages 12179--12188, 2021.

\bibitem{ranftl2020towards}
Ren{\'e} Ranftl, Katrin Lasinger, David Hafner, Konrad Schindler, and Vladlen
  Koltun.
\newblock Towards robust monocular depth estimation: Mixing datasets for
  zero-shot cross-dataset transfer.
\newblock {\em IEEE transactions on pattern analysis and machine intelligence},
  44(3):1623--1637, 2020.

\bibitem{Rottmann21}
Peter Rottmann, Thorbjörn Posewsky, Andres Milioto, Cyrill Stachniss, and Jens
  Behley.
\newblock Improving monocular depth estimation by semantic pre-training.
\newblock In {\em 2021 IEEE/RSJ International Conference on Intelligent Robots
  and Systems (IROS)}, pages 5916--5923, 2021.

\bibitem{ruder2017overview}
Sebastian Ruder.
\newblock An overview of multi-task learning in deep neural networks.
\newblock {\em arXiv preprint arXiv:1706.05098}, 2017.

\bibitem{saha2021learning}
Suman Saha, Anton Obukhov, Danda~Pani Paudel, Menelaos Kanakis, Yuhua Chen,
  Stamatios Georgoulis, and Luc Van~Gool.
\newblock Learning to relate depth and semantics for unsupervised domain
  adaptation.
\newblock In {\em Proceedings of the IEEE/CVF Conference on Computer Vision and
  Pattern Recognition}, pages 8197--8207, 2021.

\bibitem{sener2018multi}
Ozan Sener and Vladlen Koltun.
\newblock Multi-task learning as multi-objective optimization.
\newblock {\em Advances in neural information processing systems}, 31, 2018.

\bibitem{shelhamer2016fully}
Evan Shelhamer, Jonathan Long, and Trevor Darrell.
\newblock Fully convolutional networks for semantic segmentation.
\newblock In {\em Proceedings of the IEEE/CVF Conference on Computer Vision and
  Pattern Recognition}, 2016.

\bibitem{song2015sun}
Shuran Song, Samuel~P Lichtenberg, and Jianxiong Xiao.
\newblock Sun rgb-d: A rgb-d scene understanding benchmark suite.
\newblock In {\em Proceedings of the IEEE conference on computer vision and
  pattern recognition}, pages 567--576, 2015.

\bibitem{standley2020}
Trevor Standley, Amir Zamir, Dawn Chen, Leonidas Guibas, Jitendra Malik, and
  Silvio Savarese.
\newblock Which tasks should be learned together in multi-task learning?
\newblock In {\em International Conference on Machine Learning}, pages
  9120--9132. PMLR, 2020.

\bibitem{taghavi2024swinmtl}
Pardis Taghavi, Reza Langari, and Gaurav Pandey.
\newblock Swinmtl: A shared architecture for simultaneous depth estimation and
  semantic segmentation from monocular camera images.
\newblock {\em arXiv preprint arXiv:2403.10662}, 2024.

\bibitem{vandenhende2021multi}
Simon Vandenhende, Stamatios Georgoulis, Wouter Van~Gansbeke, Marc Proesmans,
  Dengxin Dai, and Luc Van~Gool.
\newblock Multi-task learning for dense prediction tasks: A survey.
\newblock {\em IEEE transactions on pattern analysis and machine intelligence},
  44(7):3614--3633, 2021.

\bibitem{DIODE}
Igor {Vasiljevic}, Nick {Kolkin}, Shanyi {Zhang}, Ruotian {Luo}, Haochen
  {Wang}, Falcon~Z. {Dai}, Andrea~F. {Daniele}, Mohammadreza {Mostajabi},
  Steven {Basart}, Matthew~R. {Walter}, and Gregory {Shakhnarovich}.
\newblock {DIODE: A Dense Indoor and Outdoor DEpth Dataset}.
\newblock {\em arXiv e-prints}, page arXiv:1908.00463, Aug. 2019.

\bibitem{vaswani2017attention}
Ashish Vaswani, Noam Shazeer, Niki Parmar, Jakob Uszkoreit, Llion Jones,
  Aidan~N Gomez, {\L}ukasz Kaiser, and Illia Polosukhin.
\newblock Attention is all you need.
\newblock {\em Advances in neural information processing systems}, 30, 2017.

\bibitem{wang2019neural}
Aria Wang, Michael Tarr, and Leila Wehbe.
\newblock Neural taskonomy: Inferring the similarity of task-derived
  representations from brain activity.
\newblock {\em Advances in Neural Information Processing Systems}, 32, 2019.

\bibitem{wang2020sdcdepth}
Lijun Wang, Jianming Zhang, Oliver Wang, Zhe Lin, and Huchuan Lu.
\newblock Sdc-depth: Semantic divide-and-conquer network for monocular depth
  estimation.
\newblock In {\em Proceedings of the IEEE/CVF Conference on Computer Vision and
  Pattern Recognition}, pages 541--550, 2020.

\bibitem{wang2021domain}
Qin Wang, Dengxin Dai, Lukas Hoyer, Luc Van~Gool, and Olga Fink.
\newblock Domain adaptive semantic segmentation with self-supervised depth
  estimation.
\newblock In {\em Proceedings of the IEEE/CVF International Conference on
  Computer Vision}, pages 8515--8525, 2021.

\bibitem{wang2023images}
Xinlong Wang, Wen Wang, Yue Cao, Chunhua Shen, and Tiejun Huang.
\newblock Images speak in images: A generalist painter for in-context visual
  learning.
\newblock In {\em Proceedings of the IEEE/CVF Conference on Computer Vision and
  Pattern Recognition}, pages 6830--6839, 2023.

\bibitem{xu2018structured}
Dan Xu, Wei Wang, Hao Tang, Hong Liu, Nicu Sebe, and Elisa Ricci.
\newblock Structured attention guided convolutional neural fields for monocular
  depth estimation.
\newblock In {\em Proceedings of the IEEE conference on computer vision and
  pattern recognition}, pages 3917--3925, 2018.

\bibitem{depthanything}
Lihe Yang, Bingyi Kang, Zilong Huang, Xiaogang Xu, Jiashi Feng, and Hengshuang
  Zhao.
\newblock Depth anything: Unleashing the power of large-scale unlabeled data.
\newblock {\em arXiv:2401.10891}, 2024.

\bibitem{yang2023polymax}
Xuan Yang, Liangzhe Yuan, Kimberly Wilber, Astuti Sharma, Xiuye Gu, Siyuan
  Qiao, Stephanie Debats, Huisheng Wang, Hartwig Adam, Mikhail Sirotenko,
  et~al.
\newblock Polymax: General dense prediction with mask transformer.
\newblock In {\em Proceedings of the IEEE/CVF Winter Conference on Applications
  of Computer Vision}, pages 1050--1061, 2024.

\bibitem{zamir2018taskonomy}
Amir~R Zamir, Alexander Sax, William Shen, Leonidas~J Guibas, Jitendra Malik,
  and Silvio Savarese.
\newblock Taskonomy: Disentangling task transfer learning.
\newblock In {\em Proceedings of the IEEE conference on computer vision and
  pattern recognition}, pages 3712--3722, 2018.

\bibitem{zhang2021survey}
Yu Zhang and Qiang Yang.
\newblock A survey on multi-task learning.
\newblock {\em IEEE Transactions on Knowledge and Data Engineering},
  34(12):5586--5609, 2021.

\bibitem{Zhou2017ADE20K}
Bolei Zhou, Hang Zhao, Xavier Puig, Sanja Fidler, Adela Barriuso, and Antonio
  Torralba.
\newblock Scene parsing through ade20k dataset.
\newblock In {\em 2017 IEEE Conference on Computer Vision and Pattern
  Recognition (CVPR)}, pages 5122--5130, 2017.

\end{thebibliography}
}

\clearpage
\appendix

\section*{Supplemental Material}

\subsection*{Reproducibility Statement}
\label{sec:reproducibility} 
All results presented in this work are reproducible using the code associated with our research. Upon acceptance, we are committed to publishing the code to facilitate transparency and enable other researchers to replicate our findings.

\subsection*{Ethics Statement}
\label{sec:ethics} 
Our research relies on publicly available datasets, ensuring transparency and reproducibility in our experiments. Additionally, for datasets obtained through agreements, such as Matterport, we adhere to the respective terms and conditions outlined in the agreements \cite{matterportAgreement}. 

\subsection*{Environmental Impact}
Recognizing the environmental impact of computational resources, we are mindful of the compute resources used in our experiments. The experiments, conducted on the aforementioned hardware setup, resulted in an estimated environmental impact of approximately 29.25 million mWh, equivalent to 0.0117 metric tons of carbon dioxide. This is comparable to the emissions from driving 5.1 miles in the average gasoline-powered passenger vehicle in the US \cite{EPA_2024}. For a comprehensive overview of the environmental impact of compute \cite{patterson2021carbon}, we refer to metrics provided by platforms like the ML CO2 Impact calculator \cite{lacoste2019quantifying}.

\section{Training Details}
\label{app:training_details}
As mentioned in Section 4 of the main paper, we adopt the DINOv2 training procedure \cite{oquab2023dinov2}. In particular, we use AdamW with initial learning rate of $0.0001$ and weight decay of $0.01$, and a cosine scheduler with linear warmup for $1/3$ of the iterations. In total we train for $38400$ steps with a batch size of 4 (2 images per 2 GPUs). For Taskonomy we use a batch size of 16 (across 8 GPUs), whereas for Matterport we train twice longer. When using our method, we duplicate the optimizer and learning rate schedulers, and scale the DPT decoder learning rates by our parameter $\alpha$.

For the experiment using the Depth Anything backbone, we adapt the Depth Anything training procedure to our scheme. In particular we use an initial learning rate of $0.000161$ and a cosine scheduler without linear warmup. We train for $38400$ steps with a batch size of 16 (2 images per 8 GPUs).

\section{Hardware Details}
\label{app:hardware_details}
In our experiments, we leveraged high-performance computing (HPC) nodes equipped with 4 NVIDIA A100 GPUs with 40GB VRAM and 48 CPU cores. For most of the experiments the training process used only 2 of the GPUs, while for some we used 8 GPUs. 

\section{Additional ablations}
\label{app:ablations}

\subsection{Single-Label Dense Classification}
\label{app:single_label}
Motivated by the promising outcomes observed with our MLDC task as an auxiliary task for Monocular Depth Estimation, we simplify the problem to Single-Label Dense Classification to investigate the viability of using straightforward classification datasets as auxiliaries for MDE. This involves extracting the dominant class for each image segmentation mask and output during training and computing the CrossEntropyLoss based on the dominant classes. As depicted in \Cref{tab:mde_mix6_alpha09_single_label}, the results for single-label and MLDC are comparable. This suggests that classifying the dominant class could potentially suffice as an auxiliary task for MDE. We encourage further exploration with additional single-label classification datasets, such as ImageNet \cite{deng2009imagenet}, to validate whether they can contribute  further improvements in MDE quality.

\begin{table*}[tb]
\sisetup{
  table-align-uncertainty=true,
  separate-uncertainty=true,
  detect-all = true,
}
\newrobustcmd{\rankfirst}{\fontseries{b}\selectfont}
\newrobustcmd{\ranksecond}{\fontshape{sl}\selectfont}
\centering
\setlength{\tabcolsep}{3pt} 
\caption{AbsRel $\times 10^4$ ($\downarrow$) scores of the best task w.r.t. the DINOv2 baseline. The last column shows the result of Single-Label Dense Classification, while others are taken from \Cref{tab:mde_mix6_alpha09}. The indoor-outdoor dominated MIX6 dataset with $\alpha=0.9$ is used for all the experiments. The best and second-best results are highlighted in \textbf{bold} and \textit{italic}, respectively.}
    \begin{tabular}{
     @{} l c *{5}{S[table-format=4.0(3),detect-weight,mode=text]} c @{} 
    }
    \toprule
    \multirow{2}{*}{\begin{tabular}[c]{@{}l@{}}MDE\\ Datasets\end{tabular}} & \multirow{2}{*}{\begin{tabular}[c]{@{}c@{}}In\\ MIX6\end{tabular}} & \multicolumn{1}{c}{} & \multicolumn{4}{c}{Aux Tasks} & \multicolumn{1}{c}{} \\ \cmidrule(lr){4-7}
                & & {DINOv2}       & {Classification}  & {Segmentation}   & {Reconstruction} & {S-Classification} & {Gain \%}\\ \hline
    NYUv2       & \xmark  &    809\pm 10   & \rankfirst  696\pm 3   &  755 \pm 6  & 838\pm 7    & \ranksecond 
 727 \pm 8  & 13.9       \\
    SUN RGBD     & \cmark &  1128 \pm 11   & \ranksecond 1024\pm 10   &  1069 \pm 9  &  1111 \pm 5   & \rankfirst 1007 \pm 11   & 10.3      \\
    Matterport3D & \xmark & 1874\pm 19   & \ranksecond  1728\pm 9   & 1793 \pm  13  &  1805\pm 6   & \rankfirst 1720 \pm 5  & 8.2      \\
    Taskonomy   & \xmark  &  1506\pm 13   & \rankfirst 1481\pm 4   &   1567\pm 12  &  1543\pm 6      & \ranksecond1491 \pm 20  & 1.7    \\
    DIODE In    & \xmark  &     3588\pm 19   &  \rankfirst  3239\pm26   &   3451\pm 47  &  3368\pm31    &   \ranksecond 3289 \pm 59  & 9.7     \\
    DIODE Out   & \xmark  &     5820\pm223   &            4965\pm162   &   5085\pm172  &  \rankfirst   4530\pm91 & \ranksecond  4926 \pm 121 & 22.2\\
    \bottomrule
    \end{tabular}%
\label{tab:mde_mix6_alpha09_single_label}
\end{table*}

\subsection{Comparison of Single and Multi Source Auxiliary Tasks} 
\label{app:single_multi}
We extend our investigation to ascertain if our approach can also be applied within a single dataset, even though this is not the primary focus of our research. For SUN RGBD we use the original provided semantic segmentation labels, whereas for the other datasets we use pseudo labels generated using the approach used in PolyMax \cite{yang2023polymax} for Taskonomy. 
Note that the quality of the pseudo-labels is not guaranteed to be high, especially when images contain objects unknown to the chosen pseudo-labeler model. This can lead to noisy labels and potentially degrade the performance. 
The results reported in \Cref{tab:mde_same_alpha09} reveal that our method can also improve the MDE quality when only using a single dataset and pre-processing for both tasks, while the quality gains for each MDE dataset are worse compared to using multiple auxiliary data sources. On one hand, this demonstrates the versatility of our method, showing that it can provide improvements in both single and multiple sources scenarios. On the other hand, these results suggest that using multiple and diverse auxiliary sources should be preferable to ensure higher quality gains.

\begin{table*}[tb]
\sisetup{
  table-align-uncertainty=true,
  separate-uncertainty=true,
  detect-all = true,
}

\newrobustcmd{\rankfirst}{\fontseries{b}\selectfont}
\newrobustcmd{\ranksecond}{\fontshape{sl}\selectfont}
\centering
\setlength{\tabcolsep}{3pt}
\caption{AbsRel $\times 10^4$ ($\downarrow$) scores on various depth datasets using different auxiliary tasks and percentage gain of the best task w.r.t. the DINOv2 baseline. For every dataset, the same dataset is used as auxiliary with $\alpha=0.9$, either with original or pseudo labels. The best and second-best results are highlighted in \textbf{bold} and \textit{italic}, respectively.}
    \begin{tabular}{
     @{} l c *{4}{S[table-format=4.0(3),detect-weight,mode=text]} c @{} 
    }
    \toprule
    \multirow{2}{*}{\begin{tabular}[c]{@{}l@{}}MDE\\ Datasets\end{tabular}} & \multirow{2}{*}{\begin{tabular}[c]{@{}c@{}}Pseudo\\ Labels\end{tabular}} & \multicolumn{1}{c}{} & \multicolumn{3}{c}{Aux Tasks -- Same Dataset Only} & \multicolumn{1}{c}{} \\ \cmidrule(lr){4-6}
                & & {DINOv2}       & {Classification}  & {Segmentation}   & {Reconstruction} & {Gain \%}\\ \hline
    NYUv2       & \cmark  &  \ranksecond  809\pm 10   & \rankfirst  762 \pm 25   &   838\pm 32  & 846\pm 29    & 5.8         \\
    SUN RGBD     & \xmark & \ranksecond 1128 \pm 11   & \rankfirst 1105 \pm 32   &  1161 \pm 7  &  1167 \pm 12  & 2         \\
    Matterport3D & \cmark & 1874\pm 19   & \rankfirst 1841\pm 21   & \ranksecond 1856\pm 17  &  1864\pm 15   & 1.8         \\
    Taskonomy   & \cmark  & \rankfirst 1506\pm 13   & \ranksecond 1576\pm 16   &   1606\pm 5  &  1607\pm 8      & -4.6      \\
    DIODE In    & \cmark  &     3588\pm 19   &  \rankfirst 3447\pm67   &   \ranksecond 3554\pm90  &    3620\pm26    & 3.1        \\
    DIODE Out   & \cmark  &     \ranksecond 5820\pm223   &            5918\pm88 & 6040\pm60   &   \rankfirst 5682\pm96  &     2.4 \\ 
    \bottomrule
    \end{tabular}%
\label{tab:mde_same_alpha09}
\end{table*}

\section{Additional qualitative results}
\label{app:qualitative}

We present additional qualitative results for each dataset to demonstrate the effectiveness of our approach. Each row should be considered individually.

\begin{figure*}[tb]
\centering
\resizebox{\textwidth}{!}{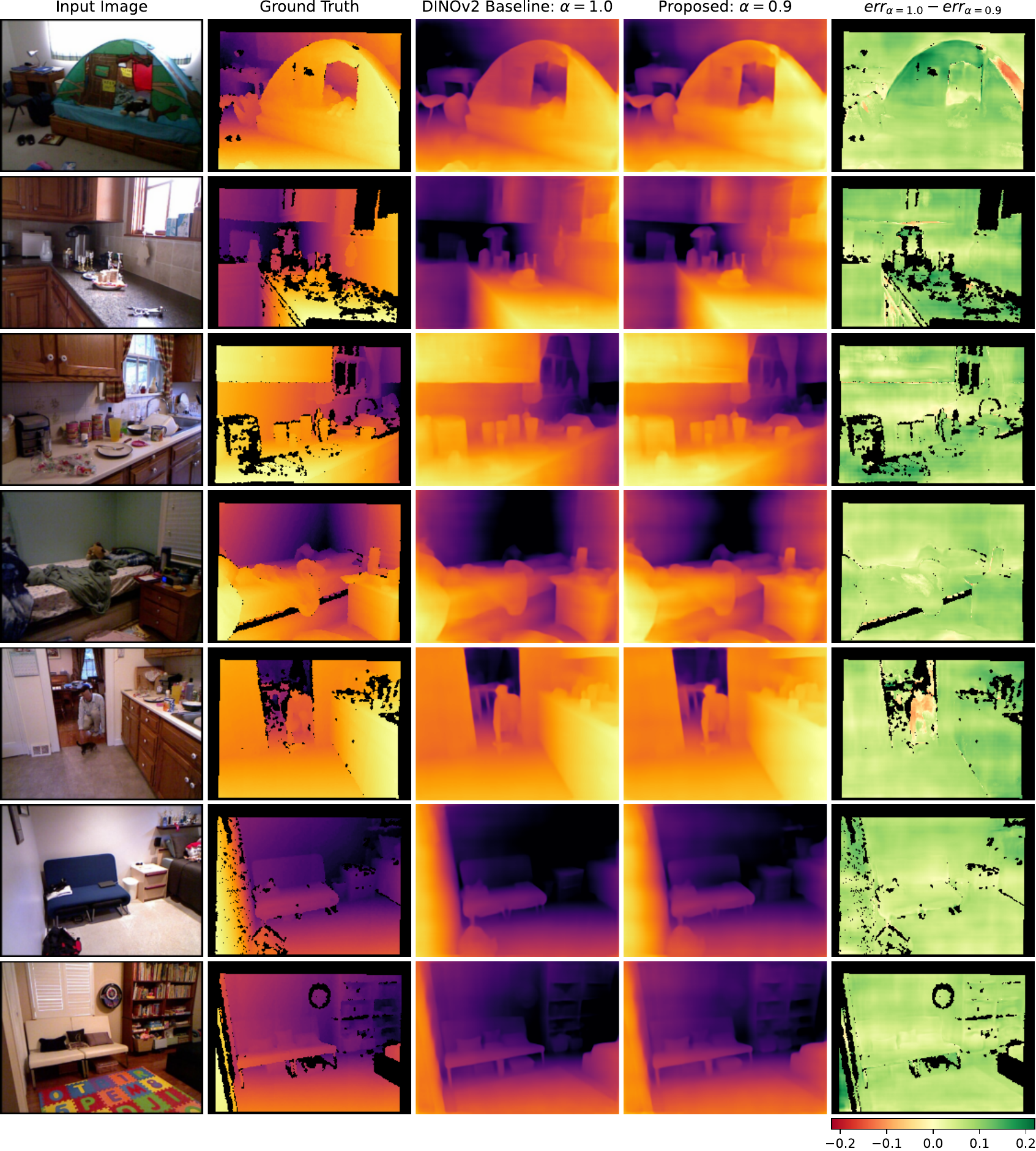}
\caption{Results on NYU with MIX6 auxiliary MLDC task. From left to right:  image and respective ground truth, baseline and our method predictions, and error difference between the last two w.r.t. to the ground truth.}
\label{fig:nyu_qualitative}
\end{figure*}

\begin{figure*}[tb]
\centering
\resizebox{\textwidth}{!}{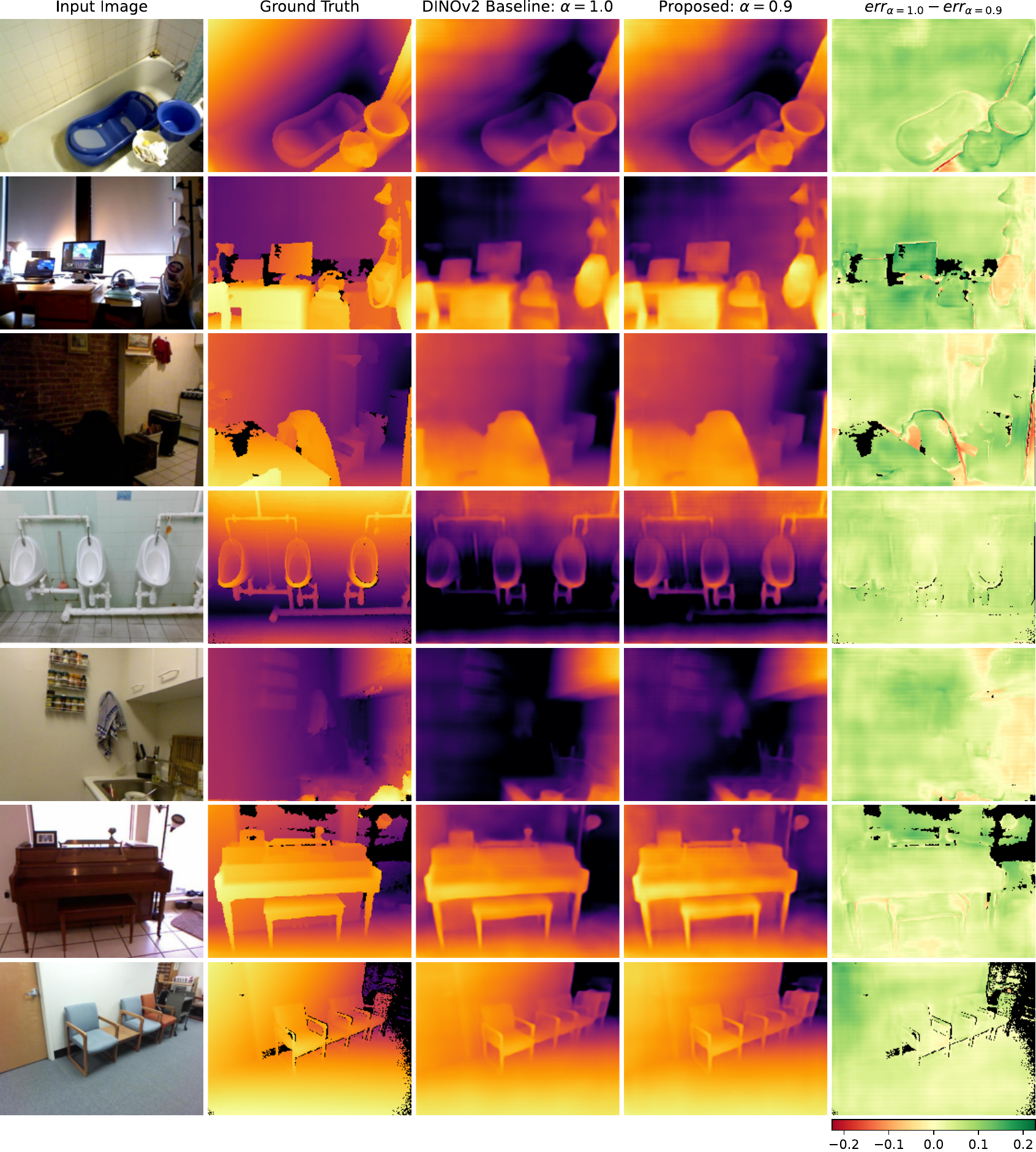}
\caption{Results on SUNRGBD with MIX6 auxiliary MLDC task. From left to right: image and respective ground truth, baseline and our method predictions, and error difference between the last two w.r.t. to the ground truth.}
\label{fig:sunrgbd_qualitative}
\end{figure*}

\begin{figure*}[tb]
\centering
\resizebox{\textwidth}{!}{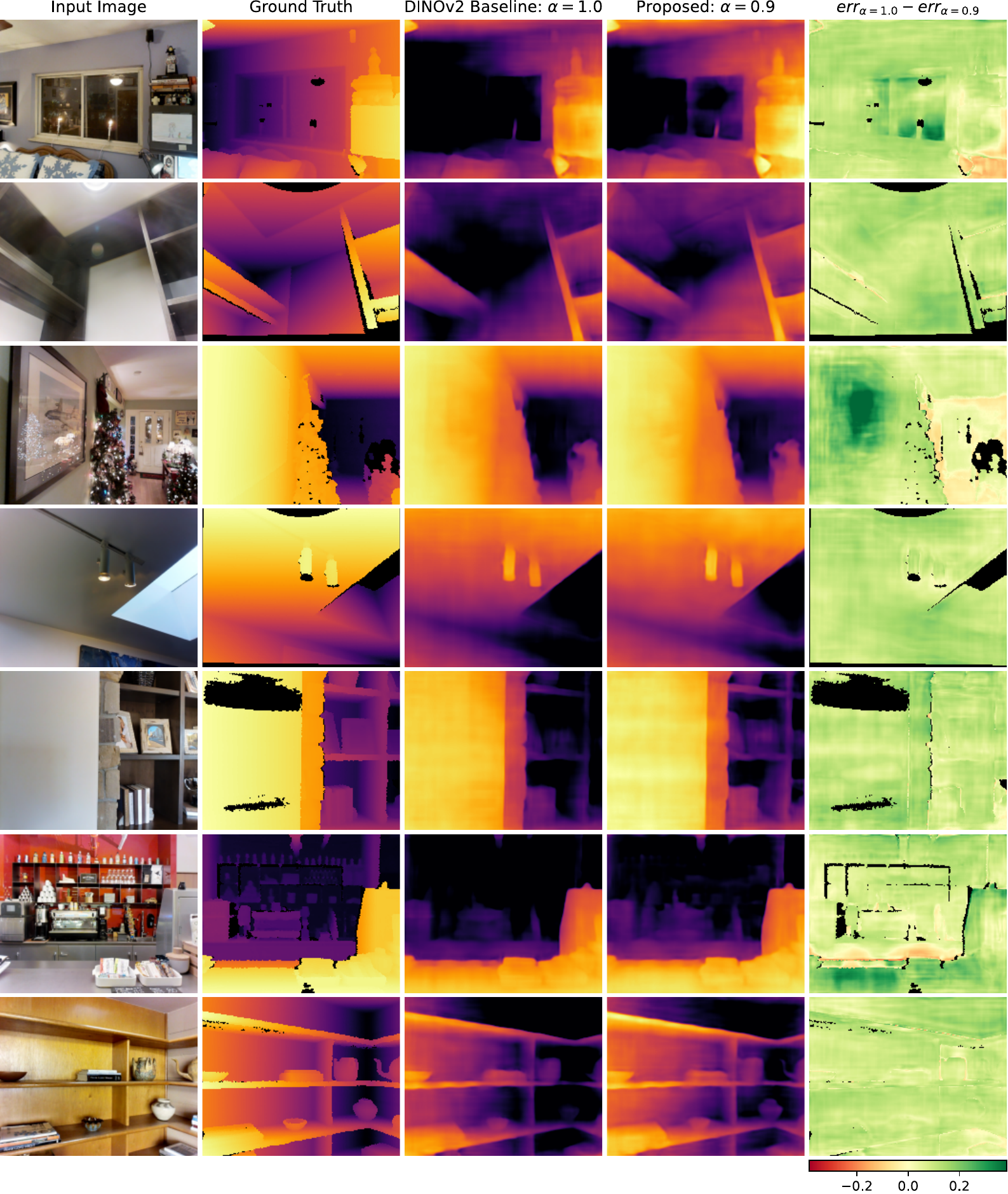}
\caption{Results on Matterport with MIX6 auxiliary MLDC task. From left to right: input image and respective ground truth, baseline and our method predictions, and error difference between the last two w.r.t. to the ground truth.}
\label{fig:matterport_qualitative}
\end{figure*}

\begin{figure*}[tb]
\centering
\resizebox{\textwidth}{!}{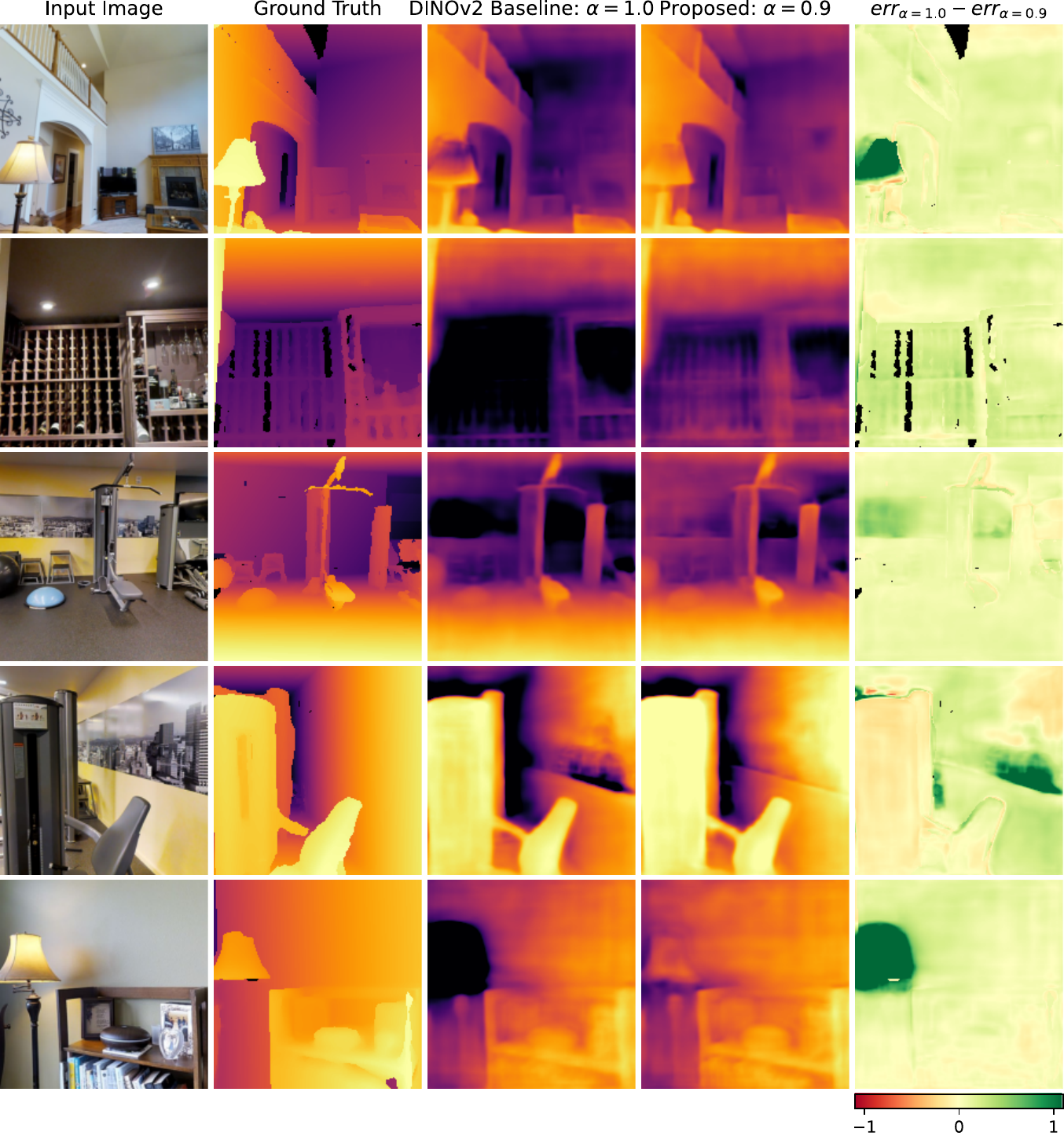}
\caption{Results on Taskonomy with MIX6 auxiliary MLDC task. From left to right: input image and respective ground truth, baseline and our method predictions, and error difference between the last two w.r.t. to the ground truth.}
\label{fig:taskonomy_qualitative}
\end{figure*}

\begin{figure*}[tb]
\centering
\resizebox{\textwidth}{!}{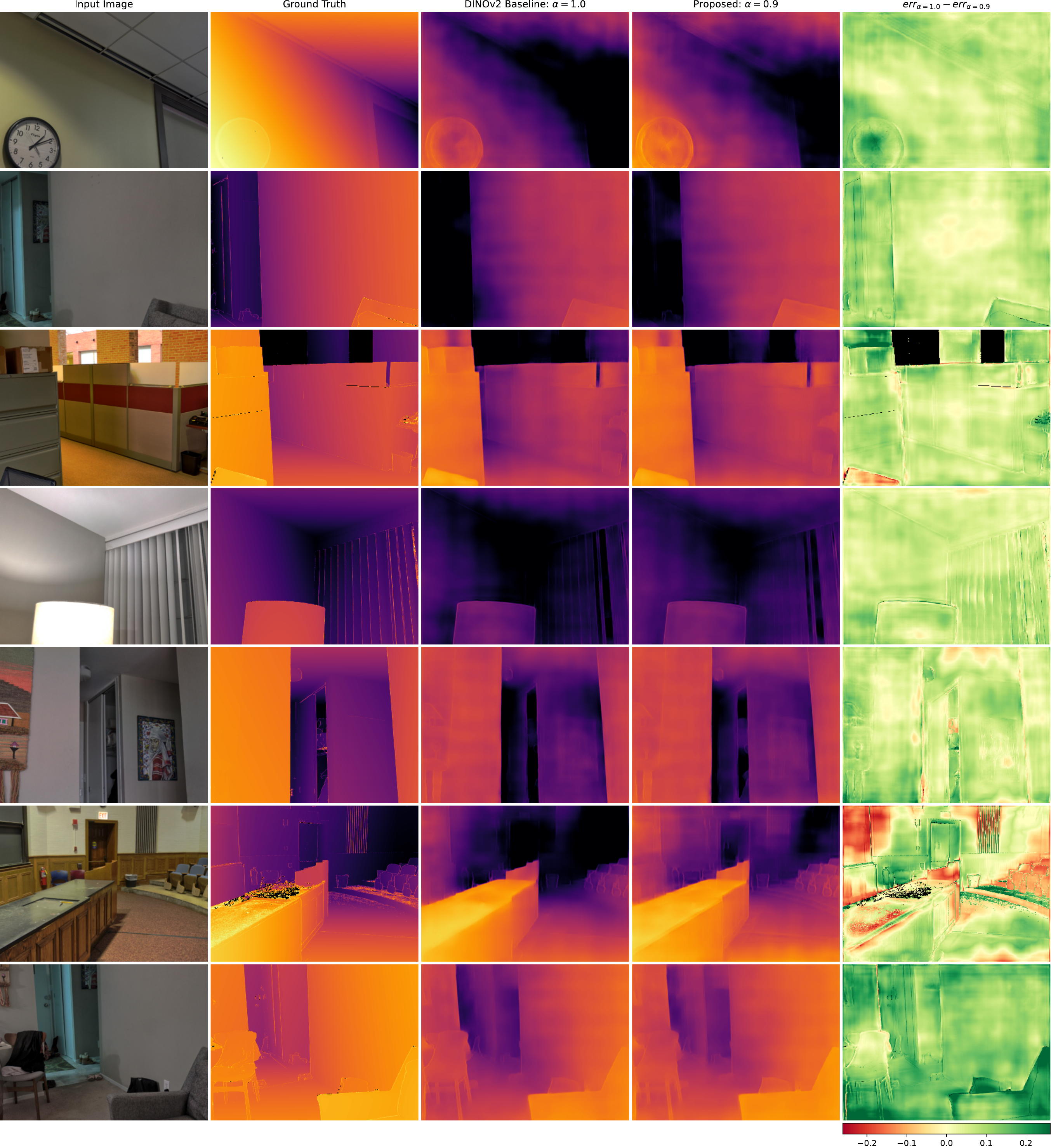}
\caption{Results on DIODE Indoor with MIX6 auxiliary MLDC task. From left to right: input image and respective ground truth, baseline and our method predictions, and error difference between the last two w.r.t. to the ground truth.}
\label{fig:diodeindoor_qualitative}
\end{figure*}

\begin{figure*}[tb]
\centering
\resizebox{\textwidth}{!}{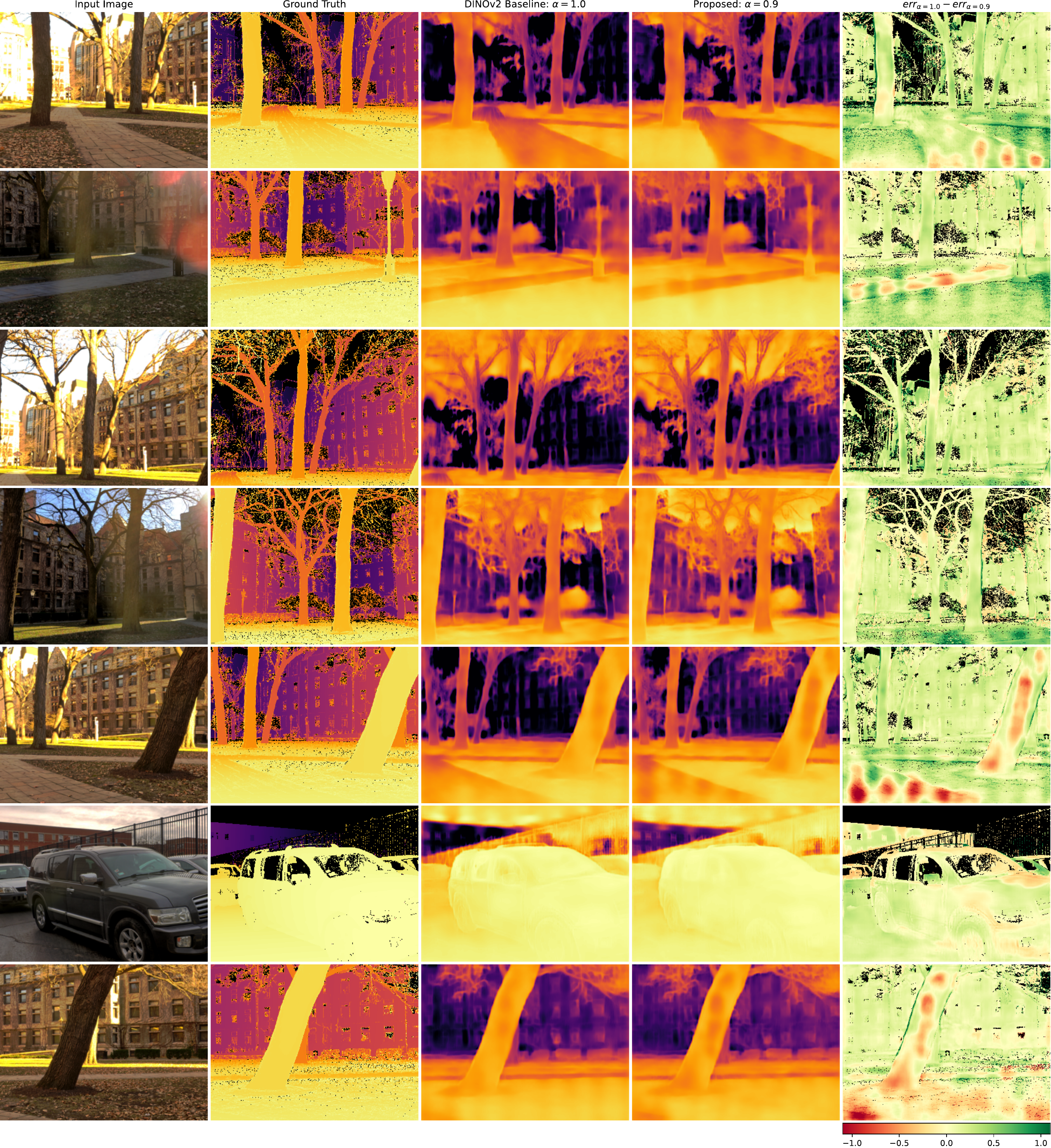}
\caption{Results on DIODE Outdoor with MIX6 auxiliary MLDC task. Left to right: input image and ground truth; baseline and our method predictions; error difference relative to ground truth highlighting visible improvements in the facade behind the trees.}
\label{fig:diodeoutdoor_qualitative}
\end{figure*}

\end{document}